\documentclass[lettersize,journal]{IEEEtran}
\usepackage{amsmath,amsfonts}
\usepackage{algorithmic}
\usepackage{algorithm}
\usepackage{array}
\usepackage[caption=false,font=normalsize,labelfont=sf,textfont=sf]{subfig}
\usepackage{textcomp}
\usepackage{stfloats}
\usepackage{url}
\usepackage{verbatim}
\usepackage{graphicx}
\usepackage{cite}
\usepackage{latexsym}
\usepackage{amsmath}%
\usepackage{MnSymbol}%
\usepackage{wasysym}%
\usepackage{booktabs}
\usepackage{enumitem}
\usepackage{graphicx}
\usepackage{color}
\usepackage{pifont}
\newcommand{\cmark}{\ding{51}}%
\newcommand{\xmark}{\ding{55}}%
\usepackage{amsthm}
\theoremstyle{definition}
\usepackage{amsfonts} 
\usepackage{wasysym}
\usepackage{tablefootnote}
\usepackage{threeparttable}
\usepackage{graphicx}
\usepackage{hhline}
\usepackage{booktabs}
\usepackage{siunitx}
\DeclareMathOperator*{\medotimes}{\text{\raisebox{0.25ex}{\scalebox{0.8}{$\bigotimes$}}}}
\DeclareMathOperator*{\medodot}{\text{\raisebox{0.25ex}{\scalebox{0.8}{$\bigodot$}}}}
\newcommand{\para}[1]{{\vspace{2pt} \bf \noindent #1 \hspace{0.5pt}}}
\usepackage{multirow}
\usepackage{dsfont}
\usepackage{colortbl}
\usepackage[numbers]{natbib}
\usepackage{caption}
\usepackage{subcaption}

\begin{document}

\title{A Probabilistic Fluctuation based Membership Inference Attack \\ for Diffusion Models}

\author{\IEEEauthorblockN{Wenjie Fu, 
Huandong~Wang,~\IEEEmembership{Member,~IEEE},
Liyuan~Zhang, 
Chen~Gao, \\
Yong~Li,~\IEEEmembership{Senior Member,~IEEE},
and~Tao~Jiang, ~\IEEEmembership{Fellow,~IEEE}} 
\thanks{Wenjie Fu and Tao Jiang are with the Research Center of 6G Mobile Communications, School of Cyber Science and Engineering, Huazhong University of Science and Technology, Wuhan 430074, P. R. China (e-mail: wjfu99@outlook.com, tao.jiang@ieee.org).}
\thanks{Huandon Wang, Chen Gao and Yong Li are with the Department of Electronic Engineering, Tsinghua University, Beijing, 100190, P. R. China (e-mail: wanghuandong@tsinghua.edu.cn, chgao96@tsinghua.edu.cn, liyong07@tsinghua.edu.cn).}
\thanks{Liyuan Zhang is with Dalian University of Technology, Dalian, 116024, P. R. China (e-mail: zhangliyuan@mail.dlut.edu.cn).}
\thanks{An arXiv version of this paper can be found in \cite{fu2023probabilistic}.}
}

\markboth{Journal of \LaTeX\ Class Files,~Vol.~14, No.~8, August~2021}%
{Shell \MakeLowercase{\textit{et al.}}: A Sample Article Using IEEEtran.cls for IEEE Journals}


\maketitle

\begin{abstract}
  Fueled with extensive and manifold training data, generative models have demonstrated extraordinary creativity in multiple fields. Nevertheless, how to avoid the misuse of privacy-sensitive and copyright-ambiguous data remains an open problem. Membership inference attacks (MIAs) offer a potential remedy by inferring whether a suspicious data record is part of the training data of a machine learning model. Although MIAs targeting conventional classification models have received substantial attention, recent research has started delving into the application of MIAs to generative models. Our investigation reveals that current MIAs tailored for generative models heavily rely on the overfitting present in the target models.
  However, overfitting can be mitigated through the application of diverse regularization techniques, resulting in subpar performance of existing MIAs in practical scenarios. In contrast to overfitting, memorization plays a crucial role in enabling deep learning models to achieve optimal performance, rendering it a more prevalent occurrence. Within generative models, memorization manifests as an upward trend in the probability distribution of generated records surrounding the member record.
  Therefore, we propose a \textbf{P}robabilistic \textbf{F}luctuation \textbf{A}ssessing \textbf{M}embership \textbf{I}nference Attack (PFAMI), a novel MIA framework designed to infer memberships by identifying the memorization pattern through an analysis of the probabilistic fluctuations surrounding specific records.
  We conduct extensive experiments across various generative models and multiple datasets, showcasing that PFAMI enhances the attack success rate (ASR) by approximately 27.9\% compared to the top-performing baseline. Our code and datasets are available in the following link\footnote{\url{https://github.com/wjfu99/MIA-Gen}}.

\end{abstract}


\section{Introduction}\label{par:intro}

In recent years, propelled by abundant data resources and empowered by deep neural networks, generative models have attained significant success in diverse domains such as computer vision \cite{rombach2022highresolution}, natural language processing \cite{li2021pretrained}, and spatial-temporal data modeling \cite{tashiro2021csdi}. These models possess the ability to produce authentic and innovative content, leading to the proliferation of various generative services \cite{stokel-walker2023what}.

While we appreciate the revolutionary benefits these services offer, we also encounter escalating privacy risks \cite{bommasani2021opportunities} and copyright disputes \cite{epstein2023art}. For instance, model privacy can be compromised by malicious users leveraging generative services \cite{hayes2018logan,zhang2020secret}. Conversely, unauthorized content might be covertly employed in developing generative services \cite{dwivedi2023opinion}.


\begin{figure}[t!]
    
    \centering
    {\includegraphics[width=.45\textwidth]{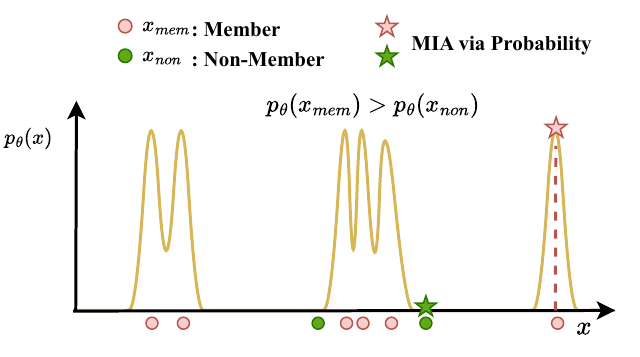}}\\
    \small (a) Overfitting\\
    {\includegraphics[width=.45\textwidth]{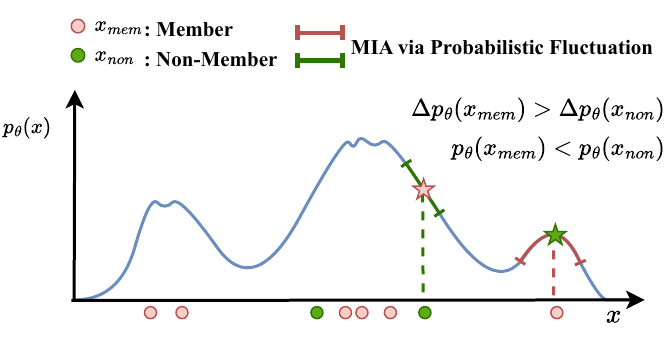}}\\
    \small (b) Memorization
    \caption{MIAs against generative models with overfitting and memorization. Identifying member records based on probability is feasible on overfitting models but fails on models only with memorization. Memorization arises as an increased tendency in probability density around member records, which can be captured by estimating the fluctuation of probability.}
    \label{fig:phenomenons}
\end{figure}

Existing works have revealed privacy leakage in generative models from multiple perspectives. Membership Inference Attacks (MIAs) are designed to ascertain whether a specific record belongs to the training set (a member record) or originates from a separate set (a non-member record) \cite{shokri2017membership}. MIAs have been widely explored in the classic classification tasks and signify significant privacy vulnerabilities \cite{hu2022membership}.
Recently, some preliminary researches have conducted MIAs on generative models \cite{chen2020ganleaks, hu2021membership, fu2023practical}. These studies primarily employ various metrics as proxies to estimate the probability of a particular record being produced by the target model, subsequently inferring memberships based on the magnitude of this probability \cite{hu2022membership}.
For example, some works \cite{hilprecht2019monte,chen2020ganleaks,liu2019performing} employ the minimum distance between generated records and the target record as a metric for approximating the aforementioned probability. Other works \cite{hayes2018logan,hu2021membership} utilize the confidence values outputted by the discriminator to serve as a proxy for this probability. 

Nevertheless, this attack paradigm, relying on the magnitude of probabilities, may implicitly necessitate model overfitting, thus undermining the efficacy of MIAs in practical scenarios.
As demonstrated in Fig.~\ref{fig:phenomenons}(a), overfitting leads member records to exhibit significantly higher probabilities of being generated compared to non-member records. Therefore, probability can serve as an indicator for membership inference.
However, training techniques such as regularization and early-stopping are widely adopted for eliminating overfitting \cite{ying2019overview}. Consequently, deep learning models are usually overfitting-free, potentially compromising the effectiveness of current attack algorithms in real-world applications.
Memorization is linked to overfitting but possesses distinct characteristics \cite{tirumala2022memorization, mireshghallah2022empirical, zhang2023counterfactual}, such as earlier occurrence \cite{tirumala2022memorization, mireshghallah2022empirical, zhang2023counterfactual}, lower detrimental impact \cite{tay2022transformer, borgeaud2022improving}, and heightened difficulty in avoidance \cite{ippolito2023preventing}. Consequently, memorization emerges as a prevalent occurrence in deep learning models and holds greater potential as a signal for membership inference.
Burg et al.~\cite{burg2021memorization} claim that the memorization in generative models should be contrasted with what is commonly associated with memorization due to overfitting. 
As shown in Fig.~\ref{fig:phenomenons}(b), memorization in generative models causes an increased tendency of probability density when the record closely resembles a member record \cite{burg2021memorization}. Therefore, we consider a more practical and promising attack framework that captures the feature of memorization by estimating the fluctuation of probability within the local scopes of data records.
Nevertheless, three challenges need to be addressed before bringing this framework to a practical MIA.
Firstly, both conventional probability-based and our proposed probabilistic fluctuation-based attack framework rely on the proxy for approximating probabilities. However, existing proxies either rely on massive synthetic records \cite{breugel2023membership} or are only applicable in white-box settings \cite{chen2020ganleaks}. Therefore, the first challenge is estimating the probability in an efficient manner without requiring massive synthetic records.
Furthermore, the probabilistic fluctuations around the target sample are highly complex within the high-dimensional data space. Specifically, the fluctuations exhibit significant variations in different directions, and the fluctuations also differ at varying distances in the same direction. Therefore, the second challenge is how to generate representative neighbor records from an appropriate direction and various distances to depict the adjacent probabilistic fluctuations.
The final challenge lies in how to design the inference function that quantifies and aggregates the probabilistic fluctuations from multiple neighbors, in order to obtain an indicator that can effectively distinguish between members and non-members.
In this paper, we propose the \textbf{P}robabilistic \textbf{F}luctuation \textbf{A}ssessing \textbf{M}embership \textbf{I}nference ($\text{PFAMI}$) composed with three elaborately designed modules to address these challenges. 
First, we design an effective probability estimation method based on variational inference, which is feasible on probabilistic generative models, such as variational autoencoders (VAEs) and diffusion models (DMs), with several synthetic records.
Furthermore, we propose a dynamic perturbation mechanism that can obtain an ensemble of representative neighbor records in the local scope by adjusting the strength of the perturbation. This ensemble paves the way for comprehensively characterizing the distribution of probabilistic fluctuations around the target record.
Finally, based on the distribution of probabilistic fluctuations, we design a statistical metric-based as well as a neural networks (NNs)-based inference function to extract the distinctive memorization features for distinguishing between members and non-members.

Overall, our contributions are summarized as follows:

\begin{itemize}[leftmargin=*]
    \item      We demonstrate that detecting memorization sheds light on MIAs against overfitting-free generative models and propose a framework that detects the distinctive characteristics of member records memorized by the target model from the perspective of probabilistic fluctuations.
    \item We propose $\text{PFAMI}$, a novel MIA method that incorporates a probability estimation approach via variational inference and a dynamic perturbation mechanism to characterize the distribution of probabilistic fluctuations, and two elaborately designed inference functions, including a metric-based approach and an NNs-based approach, to extracts essential features for membership inference attack.
    \item We conducted extensive experiments to validate the effectiveness of $\text{PFAMI}$. The results suggest that $\text{PFAMI}$ shows significantly higher ASR and stability across multiple generation models and datasets compared with existing MIAs (about $19.8\%$ and $35.9\%$ improvement in ASR on diffusion models and VAEs, respectively). 
\end{itemize}




\section{Related Works}

\begin{table*}
\centering
\caption{Taxonomy of MIAs against generative models. $\otimes$ and $\odot$ denote the w/ or w/o access requirements of specific parts. $\square$ and $\blacksquare$ indicate white-box and black-box accesses of specific parts. \cmark~ and \xmark~ represent whether an attacking algorithm is feasible for the corresponding generative model. Synthetic denotes a large-scale generated dataset prepared for MIAs in advance.}

\label{tab:taxonomy}
\resizebox{0.9\linewidth}{!}{%
\begin{tabular}{ccccccccc} 
\hline
\multirow{2}{*}{Method}        & \multicolumn{3}{c}{Access}               &  & \multicolumn{4}{c}{~Applicable}          \\ 
\cline{2-4}\cline{6-9}
                               & Generator      & Discriminator & Synthetic &  & Overfittting & Memorization & DMs & VAEs  \\ 
\hline\hline
$ \text{Co-Membership Attack}$ & $\square$      &   $\medotimes$    &    $\medotimes$      &  &    \cmark          &     \xmark        &  \cmark   &  \cmark     \\
$\text{GAN-Leaks (White-Box)}$ & $\square$      &    $\medotimes$         &    $\medotimes$       &  &      \cmark        &    \xmark         &  \cmark   &   \cmark    \\ \hline
Over-Representation (O-R)            & $\blacksquare$ &      $\medodot$       &   $\medotimes$        &  &     \cmark         &     \xmark        &  \xmark   &   \xmark    \\
LOGAN                          & $\blacksquare$ &       $\medodot$      &      $\medotimes$     &  &      \cmark        &    \xmark         &   \xmark  &   \xmark    \\ \hline
(O-R) + Surrogate Model            & $\blacksquare$ &     $\medotimes$        &   $\medodot$        &  &       \cmark       &       \xmark      &   \cmark  &    \cmark   \\
LOGAN  + Surrogate Model           & $\blacksquare$ &     $\medotimes$        &     $\medodot$      &  &     \cmark         &      \xmark       &   \cmark  &   \cmark    \\
Monte-Carlo Set                & $\blacksquare$ &      $\medotimes$       &     $\medodot$      &  &        \cmark      &      \xmark       &   \cmark  &   \cmark    \\ \hline
GAN-Leaks (Black-Box)          & $\blacksquare$ &       $\medotimes$      &     $\medotimes$      &  &        \cmark      &       \xmark      &   \cmark  &   \cmark    \\
$\text{SecMI}_\textit{Stat}$   & $\blacksquare$ &      $\medotimes$       &     $\medotimes$      &  &        \cmark      &      \xmark       &  \cmark   &    \xmark   \\
$\text{SecMI}_\textit{NNs}$    & $\blacksquare$ &       $\medotimes$      &     $\medotimes$      &  &      \cmark        &      \xmark       &   \cmark  &    \xmark   \\ 
\hline
$\text{PFAMI}_\textit{Met}$    & $\blacksquare$ &      $\medotimes$       &     $\medotimes$      &  &      \cmark        &      \cmark       &  \cmark   &   \cmark    \\
$\text{PFAMI}_\textit{NNs}$    & $\blacksquare$ &     $\medotimes$        &     $\medotimes$      &  &       \cmark       &       \cmark      &   \cmark  &   \cmark    \\
\hline
\end{tabular}
}
\end{table*}

\subsection{Generative Models}
With the development of deep learning, massive deep generative models are proposed for generating authentic data samples~\cite{oussidi2018deep}. VAEs~\cite{kingma2022autoencoding,higgins2022betavae}, a family of preliminary generative models, incorporate an encoder network to map the original data distribution into a Gaussian distribution and a decoder to generate reconstructed data from the latent distribution. 
Recently, diffusion models~\cite{ho2020denoising,song2020denoising} explore to construct desired data samples from the noise by learning a parameterized denoising process in a Markov chain. They become a new family of state-of-the-art generative models and achieve a dominant position in generative tasks such as image generation \cite{rombach2022highresolution}, text generation \cite{li2021pretrained}, and spatial-temporal data imputation \cite{tashiro2021csdi}. In this work, we elaborately evaluate the vulnerabilities of both VAEs and diffusion models to existing MIAs and PFAMI.

\subsection{Membership Inference Attack}
Shokri et al. formally proposed the MIAs, which aim to determine if a specific data record was included in the training set of a target model~\cite{shokri2017membership}. 
Recently, with the emergence of generative models, researchers have focused on exploring their vulnerabilities regarding MIA. As the taxonomy summarized in Table~\ref{tab:taxonomy}, several studies~\cite{chen2020ganleaks,liu2019performing} assume a white-box access of generators and search the smallest distance between the target record and generated records via the first-order optimization. Other works with black-box access measure this distance through a massive synthetic dataset generated by the target model \cite{hilprecht2019monte}. Additionally, there are specific studies that employ the discriminator's confidence value as a metric to differentiate member and non-member records \cite{hayes2018logan,hu2021membership}, while others adopt the estimation error \cite{duan2023are}. 
However, existing methods are only feasible on models with overfitting, and fail on models only with memorization. On the contrary, our approach infers the memberships by detecting distinctive characteristics of the training records memorized by the target model, leading to outstanding attack performance and reduced access requirements. Notably, a contemporary study proposes an MIA, neighbor attack~\cite{mattern2023membership}, for language models. While neighbor attack and PFAMI exhibit some similarities in implementations, they are fundamentally based on distinct intuitions. The PFAMI methodology is driven by the inherent phenomena in generative models, which offer profound theoretical underpinnings. Conversely, the neighbor attack suggests counteracting the bias caused by the inherent complexity of the target text by comparing it with neighboring samples.
\section{Preliminary}
In this section, we introduce representative generative models and present a formal definition of a gray-box threat model. The key notations utilized in this paper will be described in Table~\ref{tab:notation}.

\renewcommand{\arraystretch}{1.5}
\begin{table*}[h]
\footnotesize
    \begin{center}
        \caption{Notations and descriptions.}\label{tab:notation}
        \resizebox{0.9\linewidth}{!}{%
        \begin{tabular}{m{2cm}<{\centering}|m{11cm}}
            \toprule
            \textbf{Notation} & \textbf{Description} \\ \hline
             $\boldsymbol{x}^{(i)}$ & A specific data record.\\ \hline
             $\widetilde{\boldsymbol{x}}^{(i)}_j$ & A neighbor record of the target record  $\boldsymbol{x}^{(i)}$ .\\ \hline
             $m^{(i)}$ & The membership of the data record $\boldsymbol{x}^{(i)}$, 1 represents member, whereas 0 represents non-member.\\   \hline
             $\theta$ & The parameters of the target generative model.\\ \hline
             $\mathcal{A}\left( \boldsymbol{x}^{(i)}, \theta \right)$ & The adversary algorithm for MIA.\\ \hline
             $p_{\theta}\left(\boldsymbol{x}^{(i)}\right)$ & The probability of record $\boldsymbol{x}^{(i)}$ being generated by the generative model $\theta$ .\\ \hline
              $\widehat{p}_{\theta}\left(\boldsymbol{x}^{(i)}\right)$ & The approximate value of probability  $p_{\theta}\left(\boldsymbol{x}^{(i)}\right)$.\\ \hline
              $\Delta \widehat{p}_{\theta}\left(\boldsymbol{x}^{(i)}, \widetilde{\boldsymbol{x}}^{(i)}_j\right)$ & The probabilistic fluctuation between the target record $\boldsymbol{x}^{(i)}$ and one of its neighbor $\widetilde{\boldsymbol{x}}^{(i)}_j$. \\ \hline
            $\mathcal{M}(\cdot, \lambda_j)$ & The perturbation mechanism. \\ \hline
            $\left\{\lambda_j\right\}_{j=1}^M$ & The sequence of perturbation strengths. \\ \hline
             $N$ & The query times for estimating $\widehat{p}_{\theta}\left(\boldsymbol{x}^{(i)}\right)$. \\ \hline
             $M$ & The sampled number of neighbor records. \\ \hline
             $\boldsymbol{\Delta \widehat{p}_{\theta}} (\boldsymbol{x}^{(i)})$  & The $M \times N$ probabilistic fluctuation figure of the target record $\boldsymbol{x}^{(i)}$. \\
             
            \bottomrule
        \end{tabular}
        }
    \end{center}
\end{table*}
\renewcommand{\arraystretch}{1}

\subsection{Generative Models}
In this work, we focus on the probabilistic deep generative models, which include diffusion models ~\cite{ho2020denoising} and VAEs~\cite{kingma2022autoencoding},  as these are more amenable to direct analysis of the learned probability. 

VAEs has the similar structure as an autoencoder~\cite{zhai2018autoencoder}, which is composed of two modules: probabilistic encoder $q_\phi(\boldsymbol{z}\vert\boldsymbol{x})$ and decoder $ p_\theta(\boldsymbol{x}\vert\boldsymbol{z})$. 
The approximate posterior is a multivariate Gaussian distribution parameterized by encoder $\phi$:
\begin{equation}
    q_\phi(\boldsymbol{z}\vert\boldsymbol{x}) = \mathcal{N}\left(\boldsymbol{z}; \boldsymbol{\mu}_\phi\left(\boldsymbol{x}\right), \boldsymbol{\sigma}_\phi^2\left(\boldsymbol{x}\right) \mathbf{I} \right),
\end{equation}
where $ \boldsymbol{\mu}_{\phi}$ and $\boldsymbol{\sigma}_{\phi}$ are calculated by the encoding neural networks with the input of $\boldsymbol{x}$.

$ p_\theta(\boldsymbol{x}\vert\boldsymbol{z})$ is a multivariate Gaussian or Bernoulli distribution depending on the type of data.
In the image generation task, it is set to be a Gaussian distribution:

\begin{equation}
\label{equ:vae_decoder}
    p_\theta(\boldsymbol{x}\vert\boldsymbol{z}) = \mathcal{N} \left(\boldsymbol{x}; \boldsymbol{\mu}_\theta\left(\boldsymbol{z}\right), \boldsymbol{\sigma}_\theta^2\left(\boldsymbol{z}\right) \mathbf{I} \right),
\end{equation}
where $ \boldsymbol{\mu}_{\theta}$ and $\boldsymbol{\sigma}_{\theta}$ are calculated by the decoding neural networks with the input of the latent code $\boldsymbol{z}$.

Unlike VAEs, diffusion models are learned with a fixed encoding procedure. Diffusion models includes $T$ steps forward diffusion process $q\left(\boldsymbol{x}_t \mid \boldsymbol{x}_{t-1}\right)$ and reverse denoising process $p_\theta\left(\boldsymbol{x}_{t-1} \mid \boldsymbol{x}_t\right)$, which can be respectively formulated as: 
\begin{equation}
\label{equ:forward and reverse}
\begin{gathered}
q\left(\boldsymbol{x}_t \mid \boldsymbol{x}_{t-1}\right)=\mathcal{N}\left(\boldsymbol{x}_t ; \sqrt{1-\beta_t} \boldsymbol{x}_{t-1}, \beta_t \mathbf{I}\right) \\
p_\theta\left(\boldsymbol{x}_{t-1} \mid \boldsymbol{x}_t\right)=\mathcal{N}\left(\boldsymbol{x}_{t-1} ; \boldsymbol{\mu}_{\theta}\left(\boldsymbol{x}_t, t\right), \boldsymbol{\sigma}_\theta\left(\boldsymbol{x}_t, t\right)\right),
\end{gathered}
\end{equation}
where $ \left\{ \beta_t \in \left(0, 1\right) \right\}_{t=1}^{T}$ is the variance schedule. Besides, in the forward process, there is a nice property that allows sampling $\boldsymbol{x}_t$ at any arbitrary time step $t$:
\begin{equation}
\label{equ:forward}
    \boldsymbol{x}_{t}\left(\boldsymbol{x}_0, \boldsymbol{\epsilon}\right)=\sqrt{\bar{\alpha}_t} \boldsymbol{x}_0+\sqrt{1-\bar{\alpha}_t} \boldsymbol{\epsilon},
\end{equation}
where $\alpha_t = 1 - \beta_t$, $\bar{\alpha}_t = \prod_{i=1}^t \alpha_t$ and $\boldsymbol{\epsilon} \sim \mathcal{N}\left( \boldsymbol{0}, \mathbf{I} \right) $.
It is also noteworthy that the reverse probability is tractable when conditioned on $x_0$: 
\begin{equation}
\label{equ:reverse conditional}
q\left(\boldsymbol{x}_{t-1} \mid \boldsymbol{x}_t, \boldsymbol{x}_0\right)=\mathcal{N}\left(\boldsymbol{x}_{t-1} ; \tilde{\boldsymbol{\mu}}_t\left(\boldsymbol{x}_t, \boldsymbol{x}_0\right), \tilde{\beta}_t \mathbf{I}\right),
\end{equation}
where $\quad \tilde{\boldsymbol{\mu}}_t\left(\boldsymbol{x}_t, \boldsymbol{x}_0\right):=\frac{\sqrt{\bar{\alpha}_{t-1}} \beta_t}{1-\bar{\alpha}_t} \boldsymbol{x}_0+\frac{\sqrt{\alpha_t}\left(1-\bar{\alpha}_{t-1}\right)}{1-\bar{\alpha}_t} \boldsymbol{x}_t \quad$ and $\quad \tilde{\beta}_t:=\frac{1-\bar{\alpha}_{t-1}}{1-\bar{\alpha}_t} \beta_t$.

    

\subsection{Threat Model}
In this work, we consider an adversary that attempts to infer whether a specific data record was used in the training phase of the target generative model. Three mainstream attack scenarios considered in existing works are white-box~\cite{carlini2023extracting}, black-box~\cite{hilprecht2019monte}, and gray-box~\cite{duan2023are}. In the white-box scenario, the adversary has full access to the target model, including the internal parameters of target models. However, in the black-box scenario, the attacker can only receive ultimate synthetic results and lacks knowledge of the internal workings mechanism. The gray-box scenario typically necessitates access to intermediate results of target models compared to black-box attacks, which is a practical scenario where some service providers retain ownership of models' parameters while allowing users to manipulate the intermediate synthetic results~\cite{chen2020ganleaks,Pang2023BlackboxMI}, such as the latent codes in VAEs and the unfinished denoised images in diffusion models.
Therefore, we adopt the gray-box scenario in this research to evaluate our proposed method in a realistic manner. $D$ is a dataset drawn from the real data distribution, which can be partitioned into two separate subsets: $D_{mem}$ and $D_{non}$. The target model $\theta$ is trained on $D_{mem}$, and the adversary is unaware of which data records are included in the training set $D_{mem}$.
Formally, the adversary algorithm $\mathcal{A}$ is designed to predicted whether a data record $\boldsymbol{x}^{(i)} \in D$ is in the training dataset $D_{mem}$:
\begin{equation}
    \mathcal{A}\left( \boldsymbol{x}^{(i)}, \theta \right) = 
  \mathds{1}\left[
  P\left(m^{(i)}=1|\boldsymbol{x}^{(i)}, \theta\right) \geq \tau
  \right],
\end{equation}
where $m^{(i)}=1$ indicates that the record $\boldsymbol{x}^{(i)} \in D_{mem}$ , $\tau$ denotes the threshold, $\mathds{1}$ is the indicator function.



\section{Methodology}
As shown in Fig.~\ref{fig:framework}, we introduce a novel MIA framework, which infers memberships by employing probabilistic fluctuation assessment. Subsequently, we design three modules to address three challenges posed by this framework.

\subsection{Framework}

\begin{figure*}[t!]
    \centering
    {\includegraphics[width=.85\textwidth]{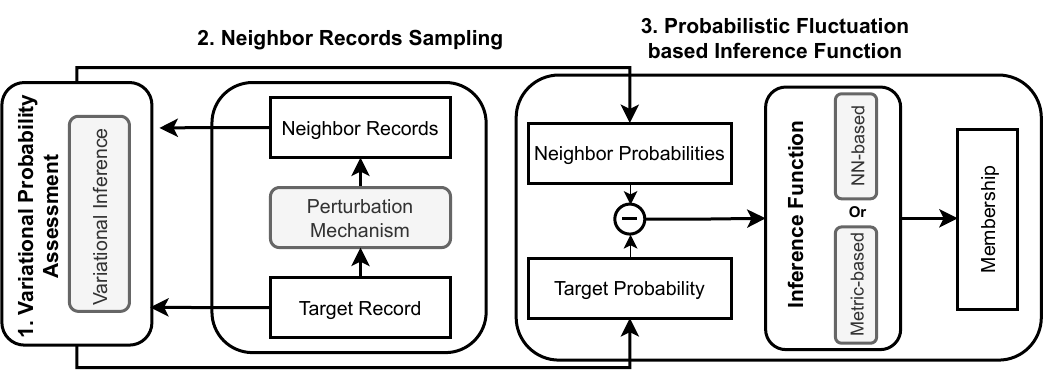}}
    \caption{The overall framework of PFAMI and the three modules introduced to deploy it in practice.}
    \label{fig:framework}
\end{figure*}

The objective of a generative model is to learn a latent variable model $p_{\theta}\left(\boldsymbol{x}\right)$ to approximate the genuine data distribution $q\left(\boldsymbol{x}\right)$. 
Thus, as we discuss in Sec.~\ref{par:intro}, there is an attack framework widely adopted by existing MIAs against generative models that approximates the probabilities of target records being generated and then sets a threshold to discriminate between member and non-member records. This framework highly relies on the overfitting phenomenon in target models that drives member records more likely to be generated than non-member records. Besides, in the training pipeline of generative models, techniques like early stopping and regularization are widely used to prevent overfitting and improve generalization \cite{ying2019overview}, which exacerbates the impracticality of this framework. Formally, the \textbf{existing attack framework} can be formulated as:
\begin{equation}
        \mathcal{A}_{exist}\left( \boldsymbol{x}^{(i)}, \theta \right) = 
  \mathds{1}\left[
  \widehat{p}_{\theta}\left(\boldsymbol{x}^{(i)}\right) \geq \tau
  \right],
\end{equation}
where $\widehat{p}_{\theta}\left(\boldsymbol{x}^{(i)}\right)$ is the approximate probability of $\boldsymbol{x}^{(i)}$ being generated.

Memorization is a phenomenon associated with, but not entirely explained by, overfitting, as evidenced by the following key distinctions: \textbf{(1) Occurrence Time}: Overfitting is typically identified as the point at which the validation loss starts to increase. In contrast, memorization occurs earlier and persists throughout almost the entire training period \cite{tirumala2022memorization, mireshghallah2022empirical, zhang2023counterfactual}. \textbf{(2) Harm Level}: While overfitting is recognized as detrimental to the generalization capabilities of machine learning models, memorization is essential for achieving near-optimal generalization in deep learning models \cite{feldman2020does}, playing a critical role in specific types of generalization tasks \cite{tay2022transformer, borgeaud2022improving}. \textbf{(3) Avoidance Difficulty}: Given that memorization occurs early in training, even with the implementation of early stopping to curb overfitting, significant memorization is likely to occur \cite{mireshghallah2022empirical}. Furthermore, as memorization proves vital for certain tasks \cite{tay2022transformer, borgeaud2022improving} and addressing specific unintended memorization patterns (such as verbatim memorization) is a challenging endeavor \cite{ippolito2023preventing}.

Thus, it is promising to infer member records based on the memorization phenomenon. In generative models, the concept of memorization contrasts with the typical association of memorization resulting from overfitting, where member records tend to yield higher generative probabilities than neighbor records \cite{burg2021memorization}, implying that it often corresponds to a local maximum.
Intuitively, we consider a more general attack framework that detects these local maximum points by leveraging the probabilistic fluctuations around target records.  Thus, \textbf{our proposed framework} can be formally represented as:

\begin{equation}
\label{equ:attack_metric}
        \mathcal{A}_{our}\left( \boldsymbol{x}^{(i)}, \theta \right) = 
  \mathds{1}\left[
    \mathcal{F} \left( \left\{ \Delta \widehat{p}_{\theta} \left(\boldsymbol{x}^{(i)}, \widetilde{\boldsymbol{x}}^{(i)}_j\right) \right\}_{j=1}^M \right) \geq \tau
  \right],
\end{equation}
where $\Delta \widehat{p}_{\theta} \left(\boldsymbol{x}^{(i)}, \widetilde{\boldsymbol{x}}^{(i)}_j\right)$ represents the probabilistic fluctuation between the target record $\boldsymbol{x}^{(i)}$  and its neighbor record $\widetilde{\boldsymbol{x}}^{(i)}_j$. $M$ indicates the number of neighbor records,  $\mathcal{F}(\cdot)$ is an inference function that can qualify the overall probabilistic fluctuation around the target record $\boldsymbol{x}^{(i)}$.

To implement our proposed framework in practical applications, we introduce three new modules sequentially. As depicted in Fig.~\ref{fig:framework}, we first propose a variational probability assessment approach to efficiently approximate the probability $\widehat{p}_{\theta}\left(\boldsymbol{x}^{(i)}\right)$ of the record $\boldsymbol{x}^{(i)}$ being generated. Then we design a dynamic perturbation mechanism to sample representative neighbors within the local scope for estimating $\Delta \widehat{p}_{\theta} \left(\boldsymbol{x}^{(i)}, \widetilde{\boldsymbol{x}}^{(i)}_j\right)$. Finally, we introduce two strategies with two inference functions to identify member records. 

\subsection{Variational Probability Assessment} \label{par:generation probability}
Existing works estimate the probability $p_\theta\left(\boldsymbol{x}\right)$ by calculating the smallest distance between the target record and the synthetic record set \cite{hu2022membership}. Nevertheless, these approaches \cite{hayes2018logan,hilprecht2019monte,breugel2023membership} rely on tens of thousands of generated records, which leads to low efficiency.

We notice that probabilistic generative models, like diffusion models and VAEs, perform training by optimizing the evidence lower bound (ELBO) of  $p_\theta\left(\boldsymbol{x}\right)$. Therefore, we attempt to derive an approximate probability $\widehat{p}_{\theta}\left(\boldsymbol{x}^{(i)}\right)$ to estimate the relative value of $p_\theta\left(\boldsymbol{x}\right)$ via variational inference.

\para{Diffusion models:} The ELBO can be written as:
\begin{equation}
 \mathds{E}_q\left[\log \frac{q\left(\boldsymbol{x}_{1: T} \mid \boldsymbol{x}_0\right)}{p_\theta\left(\boldsymbol{x}_{0: T}\right)}\right] =: L_{\mathrm{ELBO}} 
 \geq 
 -\log p_\theta\left(\boldsymbol{x}_0\right).
\end{equation}
Ho et al. further rewrite $L_{\mathrm{ELBO}}$ to $\sum_{t=0}^T L_t$, where $L_T$ is a constant can be ignored during optimization \cite{ho2020denoising}. $L_{1:T-1}$ represents the estimation error of $\boldsymbol{x}_{t-1}$, which can be expanded by applying Eq.~(\ref{equ:forward and reverse}) and Eq.~(\ref{equ:reverse conditional}):
\begin{equation}
\begin{aligned}
    L_{t-1} (\boldsymbol{x}_0) & = D_{\text{KL}}
    \left( q\left(\boldsymbol{x}_{t-1} \mid \boldsymbol{x}_t, \boldsymbol{x}_0\right) \parallel p_\theta\left(\boldsymbol{x}_{t-1} \mid \boldsymbol{x}_{t} \right) \right)\\
    & = \mathds{E}_q\left[\frac{1}{2 \sigma_t^2}\left\|\tilde{\boldsymbol{\mu}}_t\left(\boldsymbol{x}_t, \boldsymbol{x}_0\right)-\boldsymbol{\mu}_\theta\left(\boldsymbol{x}_t, t\right)\right\|^2\right], \\
\end{aligned}
\end{equation}
where the $\boldsymbol{\mu}_{\theta}$ is trained to predict $\tilde{\boldsymbol{\mu}}_t$ based on $\boldsymbol{x}_t$ and $t$.
For an attack model with the gray-box setting, it can calculate $\tilde{\boldsymbol{\mu}}_t\left(\boldsymbol{x}_t, \boldsymbol{x}_0\right)$ through the fixed forward diffusion process:
\begin{equation}
    \tilde{\boldsymbol{\mu}}_t\left(\boldsymbol{x}_t, \boldsymbol{x}_0\right) = \frac{1}{\sqrt{\alpha_t}}\left(\boldsymbol{x}_t\left(\boldsymbol{x}_0, \boldsymbol{\epsilon}\right)-\frac{\beta_t}{\sqrt{1-\bar{\alpha}_t}} \boldsymbol{\epsilon}\right),
\end{equation}
where $\boldsymbol{\epsilon} $ is sampled from $ \mathcal{N}\left( \boldsymbol{0}, \mathbf{I} \right)$, and $\boldsymbol{x}_t\left(\boldsymbol{x}_0, \boldsymbol{\epsilon}\right)$ can be calculated by following Eq.~(\ref{equ:forward}).
Then the attacker can acquire the expectation of predicted sample $\boldsymbol{\mu}_\theta\left(\boldsymbol{x}_t, t\right)$ based on Eq.~(\ref{equ:forward and reverse}) and Monte Carlo sampling:
\begin{equation}
    \boldsymbol{\mu}_\theta\left(\boldsymbol{x}_t, t\right) \approx \sum_i \boldsymbol{x}^{(i)}_{t-1},
\end{equation}
where $\boldsymbol{x}^{(i)}_{t-1} \sim p_\theta\left(\boldsymbol{x}_{t-1} \mid \boldsymbol{x}_{t} \right)$.
Thus, attacker can estimate the relative value of  $ p_\theta\left(\boldsymbol{x}_0\right)$ by fusing the estimation error across $N$ sampled time steps:
\begin{equation}
\label{equ:diffusion px}
\widehat{p}_{\theta}\left(\boldsymbol{x}_0\right) := - \frac{1}{N} \sum^N_t L_t^{\text {}} (\boldsymbol{x}_0),
\end{equation}
where $\log$ is ignored for simplification, and same operation is also performed for VAEs.

\para{VAEs:} The ELBO can be derived as the following two tractable items:
\begin{equation}
\label{equ:vae elbo}
\begin{aligned}
    - \mathds{E}_{\boldsymbol{z} \sim q_\phi(\boldsymbol{z}\vert\boldsymbol{x})} \log p_\theta(\boldsymbol{x}\vert\boldsymbol{z}) + D_\text{KL}( q_\phi(\boldsymbol{z}\vert\boldsymbol{x}) \| p_\theta(\boldsymbol{z}) ) \\
    =: L_{\mathrm{ELBO}} \geq -\log p_\theta\left(\boldsymbol{x}\right),
\end{aligned}
\end{equation}
where $\mathds{E}_{\boldsymbol{z} \sim q_\phi(\boldsymbol{z}\vert\boldsymbol{x})} \log p_\theta(\boldsymbol{x}\vert\boldsymbol{z})$ maximizes the likelihood of generating real data record $\boldsymbol{x}$, $D_\text{KL}( q_\phi(\boldsymbol{z}\vert\boldsymbol{x}) \| p_\theta(\boldsymbol{z}) )$ restricts the estimated posterior $q_\phi(\boldsymbol{z}\vert\boldsymbol{x})$ mapped from the real data record close to the prior $p_\theta(\boldsymbol{z})$.
Our adversary model is feasible to estimate both of them by only querying the target model without any knowledge of inherent parameters. 
The $\mathds{E}_{\boldsymbol{z} \sim q_\phi} \log p_\theta$ can be represented as follows by applying the Monte Carlo sampling method and Eq.~(\ref{equ:vae_decoder}):
\begin{equation}
\label{equ:vae reconstruction error}
\begin{aligned}
        \mathds{E}_{\boldsymbol{z} \sim q_\phi} \log p_\theta
        & = \frac{1}{N} \sum_{n=1}^N \log p_{\boldsymbol{\theta}}\left(\boldsymbol{x} \mid \boldsymbol{z}^{(n)}\right) \\
        & = \frac{1}{N} \sum_{n=1}^N \log \frac{1}{\boldsymbol{\sigma}_\theta \sqrt{2\pi}} e^{-\frac{1}{2}\left(\frac{\boldsymbol{x} -\boldsymbol{x}^{(n)}}{\boldsymbol{\sigma}_\theta} \right)^2} \\
        & \propto - \sum_{n=1}^N \left\| \boldsymbol{x} -\boldsymbol{x}^{(n)} \right\|^2,
\end{aligned}
\end{equation}
where $N$ represents the number of Monte Carlo sampling, $\boldsymbol{x}^{(n)}$ is the mean of the $p_\theta\left(\boldsymbol{x}\mid\boldsymbol{z}^{(n)}\right)$, i.e., $\boldsymbol{x}^{(n)}=\boldsymbol{\mu}_{\theta}^{(n)}$, and it can be directly obtained by querying the target model. Since the decoder of the VAE calculates $\boldsymbol{\mu}_{\theta} \left(\boldsymbol{z}^{(n)}\right)$ with neural networks, and outputs it as the reconstructed sample $\boldsymbol{x}^{(n)}$.
 $\left\| \boldsymbol{x} -\boldsymbol{x}^{(n)} \right\|^2$ is proportional to the mean squared error (MSE) between the original and reconstructed data records. 
$D_\text{KL}( q_\phi(\boldsymbol{z}\vert\boldsymbol{x}) \| p_\theta(\boldsymbol{z}) )$ can be calculated with a closed-form solution as it is the KL divergence between two Gaussian distribution:
\begin{equation}
\label{equ:vae KL}
D_\text{KL}( q_\phi \| p_\theta ) = \frac{1}{2} \left( 1 + \log \boldsymbol{\sigma}_\phi^2\left(\boldsymbol{x}\right) - \boldsymbol{\sigma}_\phi^2\left(\boldsymbol{x}\right) - \boldsymbol{\mu}_\phi^2\left(\boldsymbol{x}\right) \right).
\end{equation}
Thus, attacker can approximate $ p_\theta\left(\boldsymbol{x}\right)$ by applying Eq.~(\ref{equ:vae KL}) and Eq.~(\ref{equ:vae reconstruction error}) to Eq.~(\ref{equ:vae elbo}):

\begin{equation}
\label{equ:vae px}
  \widehat{p}_{\theta}\left(\boldsymbol{x}\right) :=  - \sum_{n=1}^N \left\| \boldsymbol{x} -\boldsymbol{x}^{(n)} \right\|^2 - D_\text{KL}( \mathcal{N}\left( \boldsymbol{\mu}_\phi, \boldsymbol{\sigma}_\phi^2\ \mathbf{I} \right) \| \mathcal{N}(0, 1) ).
\end{equation}

Based on Eq.~\eqref{equ:diffusion px} and  Eq.~(\ref{equ:vae px}), the adversary can approximate the relative value of $p_\theta\left(\boldsymbol{x}\right)$ by sending a few query requests to the target model.
\subsection{Neighbor Records Sampling}\label{par:perturbation}
As formulated in Eq. \eqref{equ:attack_metric}, for a given data record $\boldsymbol{x}^{(i)}$, we have to sample an ensemble of representative neighbor records $\left\{ \widetilde{\boldsymbol{x}}^{(i)}_j \right\}_{j=1}^M$. Thus, how to sample representative neighbor records for exploring the local scope of a given data record is critical for instantiating our approach. Considering neighbors approximately appressed with the target record is imperative, but the data distributions are often high-dimensional and wide. We should avoid sampling neighbors out-of-range and can therefore refrain from exploring meaningless domains. Therefore, we opt for a data perturbation method that is as simple as possible, resulting in a minor but observable shift in the data distribution. Inspired by the data augmentation techniques widely used in machine learning for improving the model performance and generalization through increasing the diversity of the dataset, we design various perturbation methods to sample neighbor records, including crop, rotation, downsampling, brightening, etc. Furthermore, we conducted extensive experiments to investigate the attack performance over various perturbation methods to find the most effective perturbation direction. The related information can be found in Appendix~\ref{app:perturbation}, and the results suggest utilizing crop as the default direction as it combines excellent performance and stability. Furthermore, we consider sampling neighbor records at different distances by adjusting the strength of perturbation to comprehensively characterize the probabilistic fluctuations around the target record.
Formally, we propose a general dynamic perturbation mechanism $\mathcal{M}$ with increasing perturbation strengths $\left\{\lambda_j\right\}_{j=1}^M$ on arbitrary perturbation method:
\begin{equation}
     \left\{ \widetilde{\boldsymbol{x}}^{(i)}_j \right\}_{j=1}^M = \left\{ \mathcal{M}\left(\boldsymbol{x}^{(i)}, \lambda_j\right) \right\}_{j=1}^M.
\end{equation}

\subsection{Probabilistic Fluctuation based Inference Function}\label{par:probabilistic fluctuation}
Based on the variational approximate probability introduced in Sec.~\ref{par:generation probability}, and the perturbation mechanism proposed in Sec.~\ref{par:perturbation}. In this section, we elaborately design two strategies with different inference functions to qualify the overall probabilistic fluctuation by analyzing the characteristics of probability changes among the neighbor records of the target record:  the metric-based inference, $\text{PFAMI}_{\textit{Met}}$, and the neural networks (NNs)-based inference,  $\text{PFAMI}_{\textit{NNs}}$.


\subsubsection{Metric-based Inference Function}
For each data record $\boldsymbol{x}^{(i)} \in D$, we sample $M$ neighbor records with increasing perturbation strengths and then set the inference function $\mathcal{F}$ to be statistical averaging to estimate the overall probabilistic fluctuation. Formally, $\text{PFAMI}_{\textit{Met}}$ can be formulated as:
\begin{equation}\label{equ: probabilistic fluctuation}
            \mathcal{A}\left(\boldsymbol{x}^{(i)}, \theta\right) = \mathds{1} \left[  \left( \frac{1}{M} \sum_{j=1}^M  \Delta \widehat{p}_{\theta} \left(\boldsymbol{x}^{(i)}, \widetilde{\boldsymbol{x}}^{(i)}_j\right)   \right) \geq \tau \right],
\end{equation}
where $\Delta \widehat{p}_{\theta} \left(\boldsymbol{x}^{(i)}, \widetilde{\boldsymbol{x}}^{(i)}_j\right) = \left( \widehat{p}(\boldsymbol{x}^{(i)}) - \widehat{p}(\widetilde{\boldsymbol{x}}^{(i)}_j) \right) / \widehat{p}(\boldsymbol{x}^{(i)})$. Besides, to provide an intuitive understanding, we demonstrate that the overall probabilistic fluctuation can be approximately interpreted as the second derivative, which is typically employed to locate critical points. Please refer to Sec.~\ref{par: interpretation} for a detailed demonstration. Note that each approximate probability $\widehat{p}_{\theta}\left(\cdot\right)$ is measured by repeatedly querying the target model $N$ times in VAE based on Eq.~(\ref{equ:vae px}). In diffusion models,  $\widehat{p}_{\theta}\left(\cdot\right)$ is calculated by taking the average of estimation errors over $N$ sampled time steps based on Eq.~(\ref{equ:diffusion px}).
\subsubsection{NNs-based Inference Function}

Instead of qualifying the overall probabilistic fluctuation by directly taking an average across $M$ neighbors and $N$ sampled points, we calculate each $\Delta \widehat{p}_{\theta}$ over them. Therefore, we can obtain a $M \times N$ matrix $\boldsymbol{\Delta \widehat{p}_{\theta}} (\boldsymbol{x}^{(i)})$ for each target record, which can be represented as an ``image" that contains probabilistic fluctuation information around the target record. Then we adopted an NNs-based model $f_{\mathcal{A}}$ as the inference function $\mathcal{F}$ to capture the information of probabilistic fluctuation variation on this image. Specifically, convolutional neural networks (CNNs)-based binary classification models are feasible to handle this task. In this manner, our proposed $\text{PFAMI}_{\textit{NNs}}$ can be formally represented as:
\begin{equation}
        \mathcal{A}\left(\boldsymbol{x}^{(i)}, \theta\right) = \mathds{1} \left[ f_{\mathcal{A}} \left(\boldsymbol{\Delta \widehat{p}_{\theta}} (\boldsymbol{x}^{(i)}) \right)  \geq \tau \right],
\end{equation}
where $f_{\mathcal{A}} (\cdot) $ indicates the probability that $\text{PFAMI}_{\textit{NNs}}$ identify target record $\boldsymbol{x}^{(i)}$ as a member.
Notably, our method does not follow existing studies \cite{duan2023are} that assume having access to an extensive number of ground truth labels for member and non-member records from the target model. Instead, we use an auxiliary dataset to train a shadow model to provide training samples for training our attack model. Then we deploy the trained attack model to infer memberships on the target model. Our method, in this manner, alleviates rigid assumptions, providing increased adaptability and practical applicability.
Clearly, the number of sampling records, $N$, and the number of neighbor records, $M$, jointly affect the total required query number $(M+1) \times N$ for the inference attack. Therefore, choosing appropriate parameters for $M$ and $N$ helps strike a balance between improving attack performance and mitigating risks that may trigger the service provider's risk management procedure. Thus, we perform experimental evaluations of the performance of PFAMI under different query numbers. For detailed information, please refer to the Sec.~\ref{par: query times}.

\section{Experiments}



\subsection{Experiment Settings}\label{par:detailed information}

\subsubsection{Datasets and Target models}
We conduct experiments on two widely-used image datasets, Celeba-64 \cite{liu2015deep} and Tiny-ImageNet (Tiny-IN) \cite{le2015tiny}. For both datasets, we randomly select about 30\% of all data samples for training and evaluating the target generative models, then utilize the remaining data for training the shadow and reference models. For example, Celeba-64 contains 202,599 images, whereas we respectively take 50,000 and 10,000 images as training and evaluation sets for target models. The detailed split and other information of the two datasets, Celeba-64 and Tiny-IN, are summarized in Table~\ref{tab:datasets}. Notably, we make every effort to use all the data samples in each dataset to ensure that the target model has sufficiently large training samples since the limited member size will exacerbate the overfitting effect. 
For the target models, we adopted the two most representative generative models, DDPM \cite{ho2020denoising} and vanilla VAE \cite{kingma2022autoencoding}, for the overall evaluation. Additionally, we also evaluated our proposed MIA against seven state-of-the-art variant diffusion models and VAEs, including DDIM~\cite{song2020denoising}, PNDM~\cite{liu2021pseudo}, LDM~\cite{rombach2022highresolution}, Beta-VAE~\cite{higgins2022betavae}, WAE~\cite{tolstikhin2018wasserstein} and RHVAE~\cite{chadebec2023data}. It is worth noting that all models employ regularization mechanisms (AdamW \cite{loshchilov2017decoupled} and early-stopping) to avoid overfitting. 



\subsubsection{Implementation Details}

\begin{table*}[t]
    \centering
    \caption{Detailed split and other information of datasets.}
    \label{tab:datasets}
    \resizebox{\linewidth}{!}{%
    \begin{tabular}{c|clcclcclccl} 
    \cline{1-11}
    \multirow{2}{*}{Dataset} & \multirow{2}{*}{Resolution} & \multirow{2}{*}{} & \multicolumn{2}{c}{Target Model} &  & \multicolumn{2}{c}{Shadow Model} &  & \multicolumn{2}{c}{Reference Model} &   \\ 
    \cline{4-5}\cline{7-8}\cline{10-11}
                             &                             &                   & \# Member & \# Non-member        &  & \# Member & \# Non-member        &  & \# Member & \# Non-member           &   \\ 
    \hhline{===========~}
    Celeba-64                & 64                          &                   & 50,000    & 10,000               &  & 50,000    & 10,000               &  & 50,000    & 10,000                  &   \\
    Tiny-IN                  & 64                          &                   & 30,000    & 5,000                &  & 30,000    & 5,000                &  & 30,000    & 5,000                   &   \\
    \cline{1-11}
    \end{tabular}
    }
\end{table*}

The target models are all prepared with the two most popular generative model libraries: diffusers~\cite{von-platen-etal-2022-diffusers} and pythae~\cite{chadebec2022pythae}, which allow researchers to easily deploy our attack model on other generative models with just a few lines of code modification. All diffusion models are deployed with diffusers, and trained in 500 epochs with a learning rate of 0.0001 and batch size of 16. All VAE models are deployed with pythae, and trained in 100 epochs with a learning rate of 0.0001 and batch size of 100. Note that the training process will be interrupted early if the loss starts to increase in the evaluation set. 
All target models are trained with general settings, and the backbone of diffusion models and VAEs are selected to UNet \cite{ronneberger2015u} and ResNet \cite{he2016deep}. 
The step length of diffusion models is set to $T=1,000$, and the dimension of latent variables $\boldsymbol{z}$ in VAEs is set to $64$.
To assure the target models are well-generalized, we use AdamW \cite{loshchilov2017decoupled} to optimize all generative models, which fuses the Adam optimizer \cite{kingma2014adam} and the L2 regularization to reduce the risk of model overfitting. Furthermore, we adopt early-stopping during the training process,  where we stop training before the loss increases in the validation set. The crop 
 is adopted as the default perturbation mechanism for all experiments. For $\text{PFAMI}_{\textit{Met}}$, we set $N=10$ and $M=1$ for attacking against both diffusion models and VAEs. In diffusion models, we sample time steps starting from $0$ to $500$ with an interval of $50$, as we found that the probabilistic fluctuations of member and non-member are indistinguishable in larger time steps, where the image is to Gaussian noise. To strike a more accurate probabilistic fluctuation estimation, we adopt difficulty calibration~\cite{watson2022importance} and train a reference generative model with another relevant but disjoint dataset to calibrate the approximate probability $\widehat{p}_{\theta}\left(\boldsymbol{x}_0\right)$ in both metric-based and NNs-based strategies. As for $\text{PFAMI}_{\textit{NNs}}$, we designed a set of equally spaced increasing perturbation strengths, the factor $\lambda$ ranging from $0.98$ to $0.7$, with a length of $M=10$. We choose the ResNet \cite{he2016deep} as the backbone of attack model $f_{\mathcal{A}}$ and train it with only 2,000 samples provided by the shadow model.

\subsubsection{Baselines}
We choose six state-of-the-art MIAs designed for generative models across all adversary scenarios to evaluate our proposed method comprehensively. There are two baselines set up in the white-box setting: Co-Membership \cite{liu2019performing} and GAN-Leaks (White-Box) \cite{chen2020ganleaks}. Additionally, four baseline methods are employed for the black-box access scenario: LOGAN \cite{hayes2018logan}, Monte-Carlo Set \cite{hilprecht2019monte}, Over-Representation \cite{hu2021membership}, and GAN-Leaks (Black-box) \cite{chen2020ganleaks}. Besides, SecMI \cite{duan2023are} represents the attack under gray-box access, which shares exactly the same setting as our proposed method. These baselines have been comprehensively verified to have appreciable attack performances over multifarious generative models and across myriad datasets.



\subsection{Attack Performance}

\begin{table*}[t]
    \centering
    \tabcolsep=3pt
    
    \caption{Performance of $\text{PFAMI}$ across two generative models and two datasets. $\uparrow$ represents that the higher the metric, the better of performance. \textbf{Bold} and \underline{Underline} respectively denote the best and the second-best results for each metric. \#~Query indicates the number of query requests issued to the target model by each attack algorithm to complete an attack. N/A demonstrates that $\text{SecMI}_\textit{Stat}$ and $\text{SecMI}_\textit{NNs}$ are unavailable on VAEs, since they are specially designed for diffusion models.}
    \label{tab:attack performance}
    \resizebox{0.9\linewidth}{!}{%
    
    \begin{tabular}{l|ccccccccccccc} 
    \hline
    \multirow{3}{*}{Attack Types} & \multirow{3}{*}{Method}        &          & \multicolumn{5}{c}{DDPM}                                                   &  & \multicolumn{5}{c}{VAE}                                                                               \\ 
    \cline{4-8}\cline{10-14}
                                  &                                &          & \multicolumn{2}{c}{Celeba-64}    &  & \multicolumn{2}{c}{Tiny-IN}                     &  & \multicolumn{2}{c}{Celeba-64}                   &  & \multicolumn{2}{c}{Tiny-IN}                      \\ 
    \cline{4-5}\cline{7-8}\cline{10-11}\cline{13-14}
                                  &                                & \# Query & ASR$\uparrow$           & AUC$\uparrow$   &  & ASR$\uparrow$ & AUC$\uparrow$ &  & ASR$\uparrow$ & AUC$\uparrow$ &  & ASR$\uparrow$ & AUC$\uparrow$  \\ 
    \hline\hline
    \multirow{2}{*}{White-Box}    & $ \text{Co-Membership Attack}$ & 1000     & 0.637          & 0.682           &  & 0.632                  & 0.679                  &  & 0.640                  & 0.691                  &  & 0.637                  & 0.690                   \\
                                  & $\text{GAN-Leaks (White-Box)}$ & 1000     & 0.623          & 0.601           &  & 0.618                  & 0.593                  &  & 0.601                  & 0.586                  &  & 0.595                  & 0.582                   \\ 
    \hline
    \multirow{4}{*}{Black-Box}    & Monte-Carlo Set                & 10000    & 0.500          & 0.502           &  & 0.500                  & 0.501                  &  & 0.501                  & 0.502                  &  & 0.501                  & 0.501                   \\
                                  & Over-Representation            & 10000    & 0.511          & 0.517           &  & 0.509                  & 0.510                  &  & 0.508                  & 0.514                  &  & 0.506                  & 0.513                   \\
                                  & LOGAN                          & 10000    & 0.509          & 0.507           &  & 0.508                  & 0.507                  &  & 0.505                  & 0.506                  &  & 0.506                  & 0.509                   \\
                                  & GAN-Leaks (Black-Box)          & 10000    & 0.503          & 0.505           &  & 0.502                  & 0.504                  &  & 0.505                  & 0.506                  &  & 0.502                  & 0.505                   \\ \hline
    \multirow{4}{*}{Gray-Box}    & $\text{SecMI}_\textit{Stat}$   & 12   & 0.690          & 0.741           &  & 0.673                  & 0.729                  &  & N/A                    & N/A                    &  & N/A                    & N/A                     \\
                                  & $\text{SecMI}_\textit{NNs}$    & 12   & 0.791          & 0.867           &  & 0.783                  & 0.859                  &  & N/A                    & N/A                    &  & N/A                    & N/A                     \\ 
    \cline{2-14}
                                  & $\text{PFAMI}_\textit{Met}$    & 20 & \underline{0.909}  & \underline{0.965} &  & \underline{0.900}          & \underline{0.961}        &  & \underline{0.822}         & \underline{ 0.900}         &  & \underline{0.811}          & \underline{0.893}         \\
                                  & $\text{PFAMI}_\textit{NNs}$    & 110 & \textbf{0.947} & \textbf{0.986}  &  & \textbf{0.939}         & \textbf{0.978}         &  & \textbf{0.863}         & \textbf{0.939}         &  & \textbf{0.849}         & \textbf{0.927}          \\
    \hline
    \end{tabular}
    }
\end{table*}

\begin{figure*}[h]
    \centering
    \begin{tabular}{cccc}
    {\includegraphics[width=.225\textwidth]{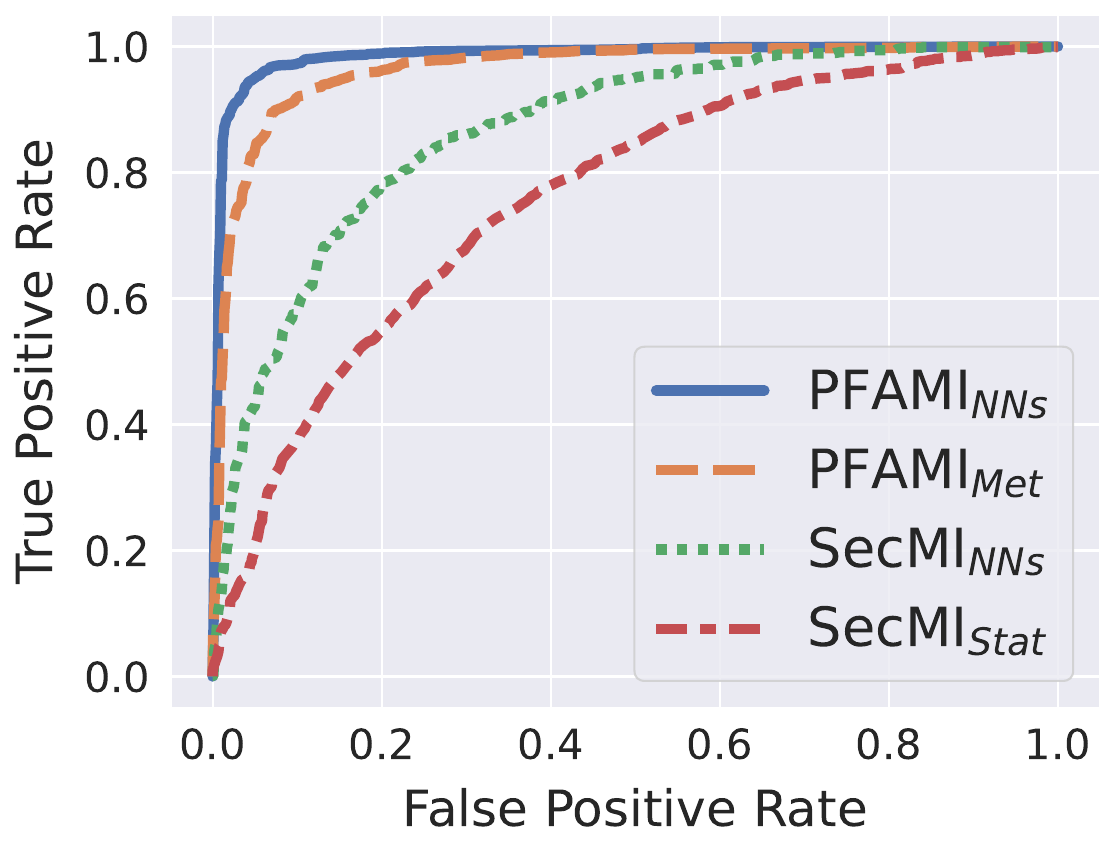}} &
    {\includegraphics[width=.225\textwidth]{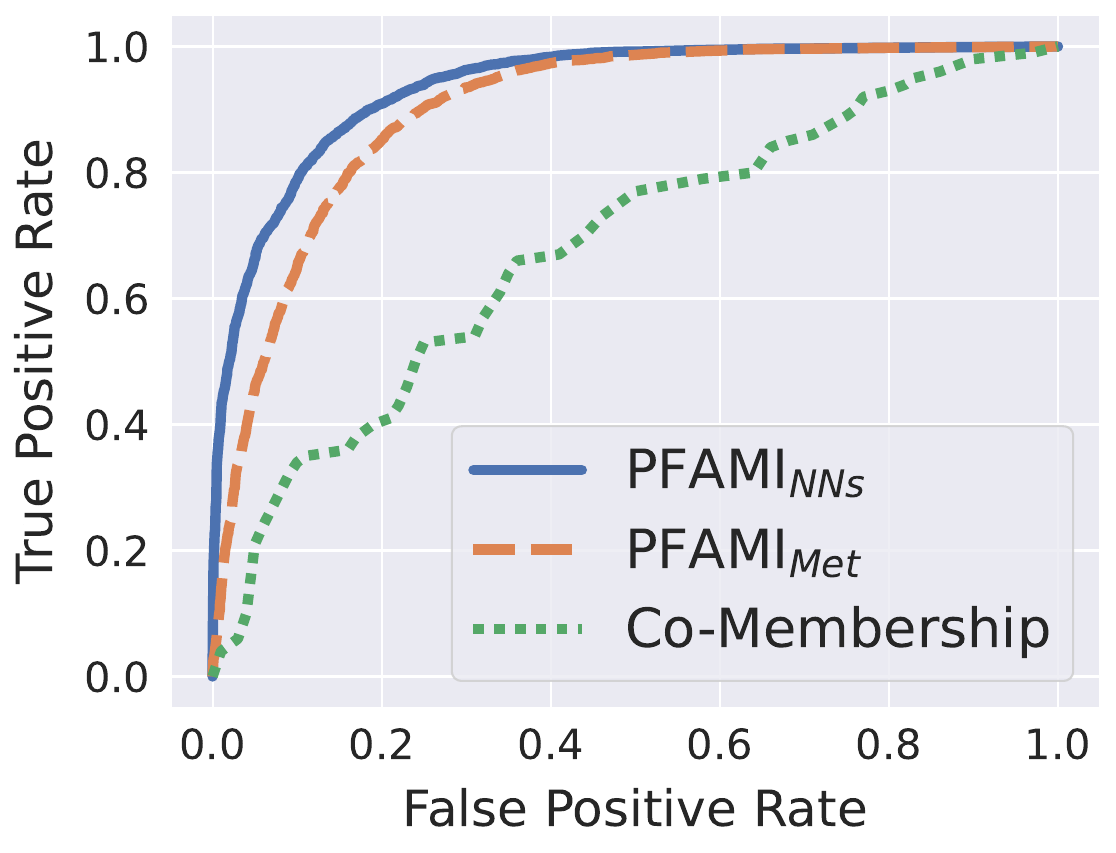}} &
    {\includegraphics[width=.225\textwidth]{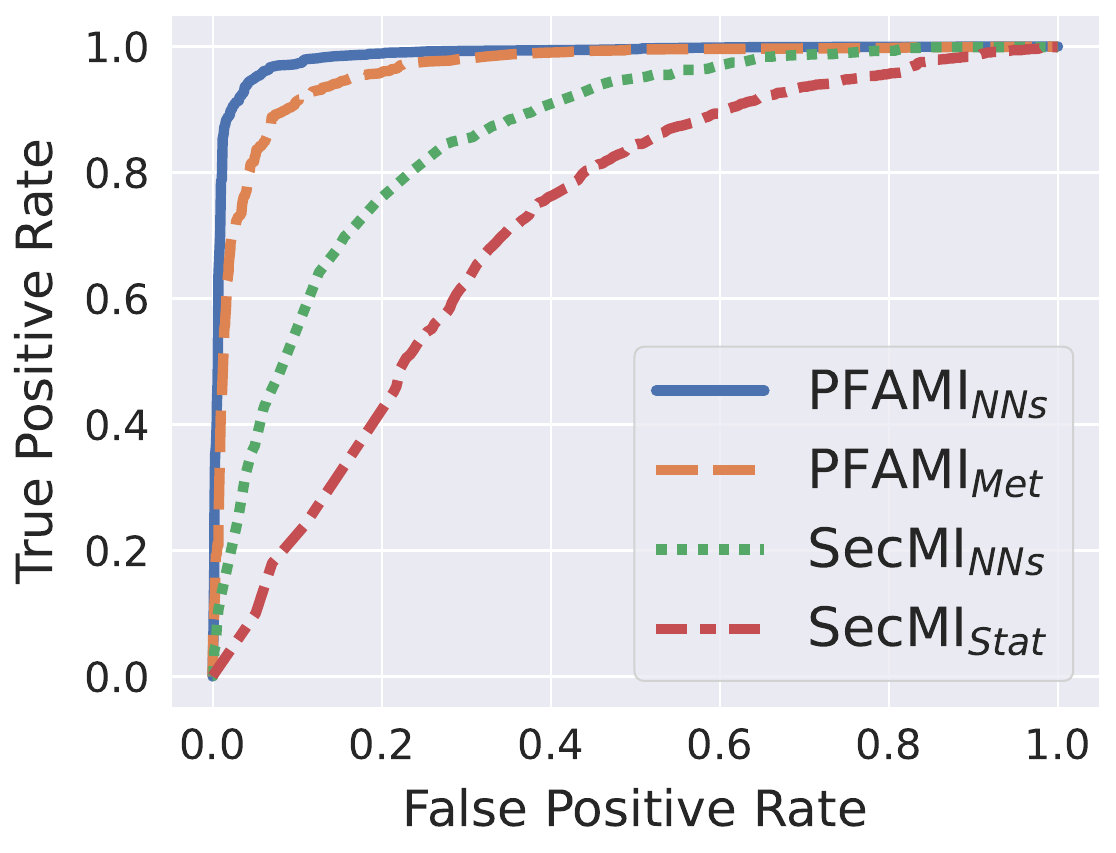}} &
    {\includegraphics[width=.225\textwidth]{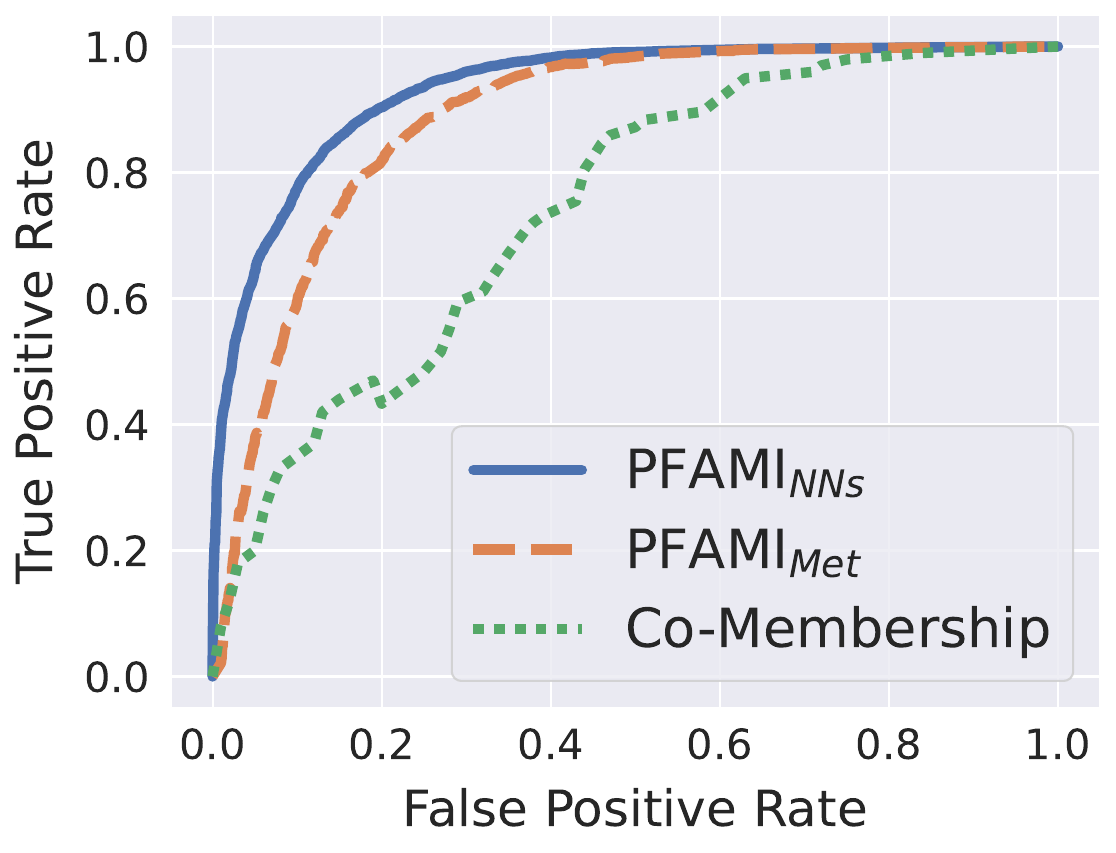}} \\
    (a) DDPM @ Celeba-64 & (b) VAE @ Celeba-64 & (c) DDPM @ Tiny-IN & (d) VAE @ Tiny-IN \\
    \end{tabular}
    \caption{ROC curves of $\text{PFAMI}$ and the best baselines on two generative models trained in Celeba-64 and Tiny-IN datasets.}\label{fig:ROC curve}
\end{figure*}

As shown in Table~\ref{tab:attack performance}, we first summarize the AUC \cite{bradley1997use} and ASR \cite{duan2023are} metrics for all baselines and $\text{PFAMI}$ against two generative models over two datasets. We also illustrate receiver operating characteristic (ROC) curves for $\text{PFAMI}$ and the best baseline in Fig.~\ref{fig:ROC curve} for a more comprehensible presentation. 
The results demonstrate that $\text{PFAMI}_{\textit{NNs}}$ and $\text{PFAMI}_{\textit{Met}}$ models achieve the highest average ASRs of $90.0\%$ and $86.1\%$ respectively. $\text{PFAMI}$ exhibits approximate $35.9\%$ improvement in AUC compared to the most competitive baseline on VAE, even when the baseline is set up with white-box access. Moreover, $\text{PFAMI}_\textit{NNs}$ achieves higher ASR and AUC in DDPM compared to $\text{SecMI}_\textit{NNs}$ that assumes access to a large number of member and non-member data records of target models. $\text{PFAMI}_\textit{NNs}$ can further increase the inference performance via capturing the variation of probabilistic fluctuations with neural networks  (about $5\%$ and  $3\%$ improvement in ASR and AUC.). Besides, both $\text{PFAMI}_{\textit{NNs}}$ and $\text{PFAMI}_{\textit{Met}}$ exhibit superior attack performance on diffusion model. This phenomenon can be attributed to the enhanced confidence in inferring membership through the estimation of probability fluctuations across different time steps in diffusion models.

\subsection{Generalizability Study}

\begin{table}[t]
\centering
\tabcolsep=0.5em
\caption{$\text{PFAMI}$ is compared with the best baseline in terms of performance against various generative models, where $\text{SecMI}_\textit{NNs}$ and Co-Membership Attack stand as the best baselines for diffusion models and VAEs. Models are trained over the Celeba-64 dataset.}
\label{tab:generalizability}
\resizebox{0.9\linewidth}{!}{%


\begin{tabular}{ccccccccc} 
\hline
\multirow{2}{*}{Target Model} & \multicolumn{2}{c}{$\text{Best Baseline}$} &  & \multicolumn{2}{c}{$\text{PFAMI}_\textit{Met}$} &  & \multicolumn{2}{c}{$\text{PFAMI}_\textit{NNs}$}  \\ 
\cline{2-3}\cline{5-6}\cline{8-9}
                              & ASR$\uparrow$ & AUC$\uparrow$              &  & ASR$\uparrow$ & AUC$\uparrow$                   &  & ASR$\uparrow$ & AUC$\uparrow$                    \\ 
\hline\hline
DDIM                          & 0.799         & 0.871                      &  & 0.913         & 0.969                           &  & 0.952         & 0.987                            \\
PNDM                          & 0.787         & 0.865                      &  & 0.907         & 0.968                           &  & 0.941         & 0.977                            \\
LDM                           & 0.786
& 0.862                      &  & 0.901         & 0.961                           &  & 0.932         & 0.968                            \\ 
\hline
Beta-VAE                      & 0.642         & 0.695                      &  & 0.824         & 0.903                           &  & 0.863         & 0.940                            \\
WAE                           & 0.667         & 0.723                      &  & 0.831         & 0.916                           &  & 0.867         & 0.944                            \\
RHVAE                         & 0.652         & 0.702                      &  & 0.817         & 0.899                           &  & 0.856         & 0.927                            \\
\hline
\end{tabular}
}
\end{table}

To validate the generalizability and robustness of our proposed attack method on probabilistic generative models, we consider six different generative models: DDIM~\cite{song2020denoising}, PNDM~\cite{liu2021pseudo} and LDM~\cite{rombach2022highresolution} as the variants of diffusion models. Beta-VAE~\cite{higgins2022betavae}, WAE~\cite{tolstikhin2018wasserstein} and RHVAE~\cite{chadebec2023data} as the variants of VAEs. As the results demonstrated in Table~\ref{tab:generalizability}, our proposed method can achieve generally well attack performance across various generative models. Furthermore, the experimental results once again demonstrate the higher privacy exposure risk of diffusion models regarding MIAs. Additionally, we also evaluate PFAMI and the most aggressive baselines on Stable Diffusion~\cite{rombach2022highresolution}, one of the most prevalent and state-of-the-art conditional diffusion models. In contrast to unconditional diffusion models and VAEs that are trained from scratch, we follow the same setup as Duan et al.~\cite{duan2023are} to fine-tune the Stable Diffusion over the Pokemon dataset. We adopt the Stable Diffusion-V1.4 released in the HuggingFace~\footnote{\url{https://huggingface.co/CompVis/stable-diffusion-v1-4}} as the pre-trained model.
The results are summarized in Table~\ref{tab:stable diffusion}, where PFAMI still strikes notable inference performances.

\begin{table}[t]
\centering
\caption{Inference Performance of $\text{PFAMI}_\textit{Met}$ and the most aggressive baselines, $\text{SecMI}_\textit{Stat}$, on Stable Diffusion.}
\label{tab:stable diffusion}
\resizebox{0.8\linewidth}{!}{%
\begin{tabular}{c|ccc} 
\hline
Method    & ASR$\uparrow$ & AUC$\uparrow$ & TPR@1\%FPR$\uparrow$  \\ 
\hline
$\text{PFAMI}_\textit{Met}$  & 0.967         & 0.993         & 0.774                 \\ 
\hline
$\text{SecMI}_\textit{Stat}$ & 0.817         & 0.886         & 0.145                 \\
\hline
\end{tabular}
}
\end{table}

\subsection{Impact of Memorization Degree on Attack Performance}

\begin{figure}[ht]
        \centering
        {\hspace{-13pt}\includegraphics[width=.47\textwidth]{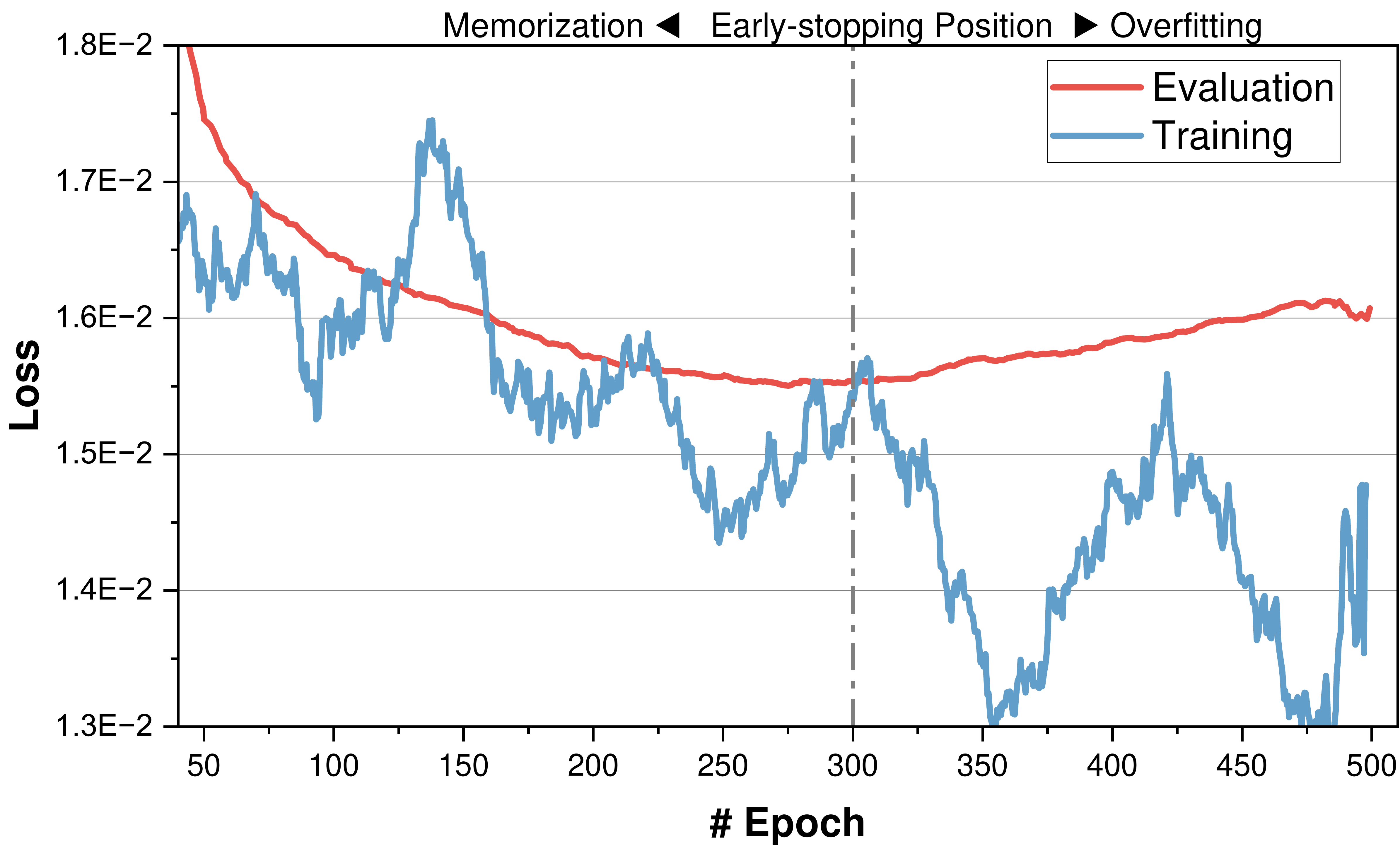}}
        \caption{The loss trajectory of DDPM@Celeba-64 on evaluation and training datasets.}\label{fig:loss_traj}
\end{figure}

\begin{figure}[ht]
        \centering
        {\includegraphics[width=.47\textwidth]{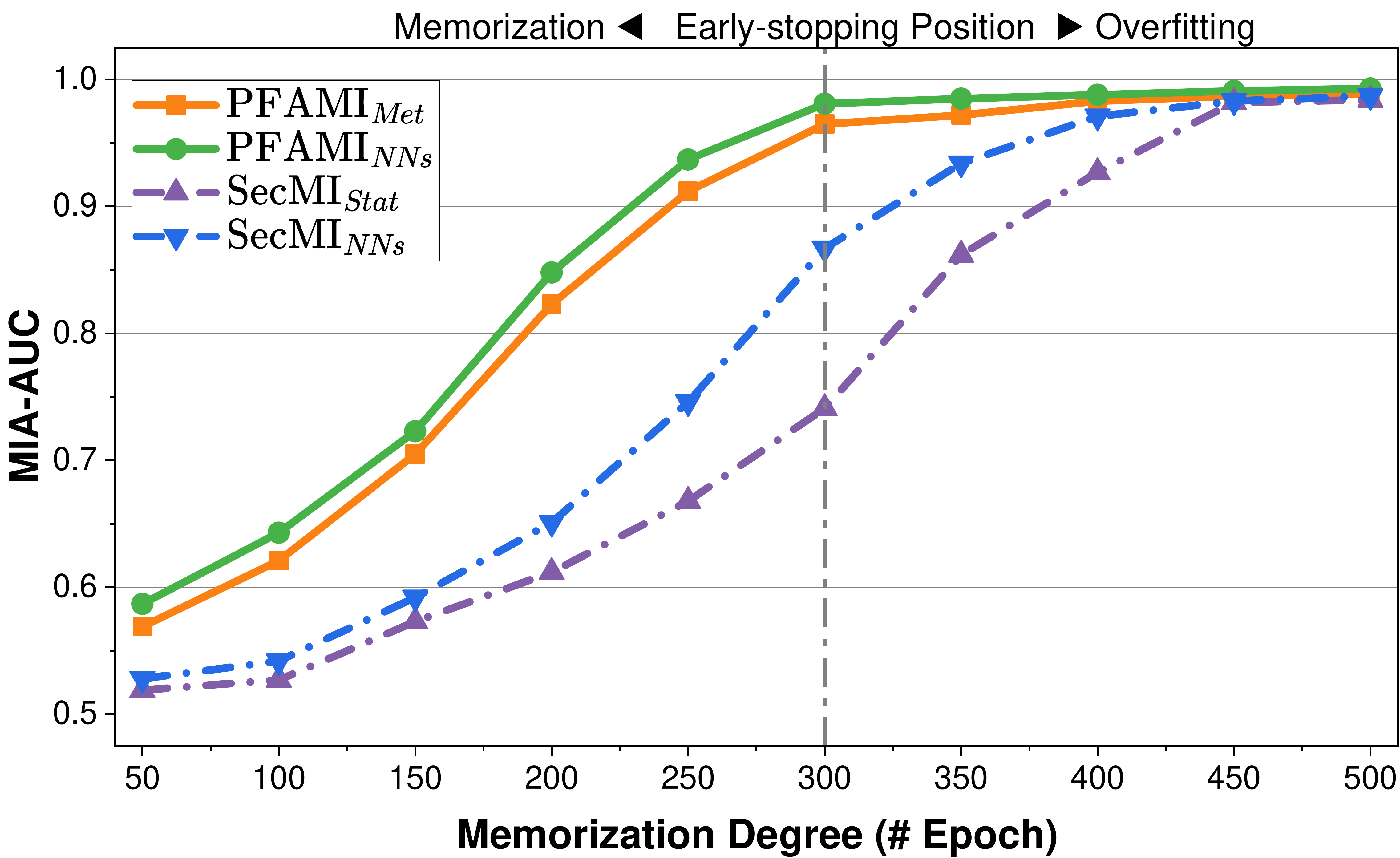}}
        \caption{The performance (AUC) of PFAMI and SecMI against DDPM under different memorization degrees.}\label{fig:mia_traj}
\end{figure}

The design of the proposed method, PFAMI, draws inspiration from the memorization characteristic of generative models, which is more prominent in overfitting-free models. Thus, it is essential to investigate the attack performance of PFAMI and the most competitive baseline on both memorization-only and overfitting generative models, each exhibiting varying degrees of memorization. We take a case study on DDPM trained in the Celeba-64 dataset, where we employ the epoch number as the proxy of memorization since the memorization degree persists increasing throughout almost the entire training phase~\cite{burg2021memorization}. 
In contrast to memorization, the occurrence of overfitting is defined to be the first epoch when the loss in the evaluation set starts to increase~\cite{tirumala2022memorization}. As shown in Fig.~\ref{fig:loss_traj}, we record the loss values of target models at each checkpoint on both training and evaluation sets. The 300th epoch is considered as the position where early-stopping will intervene to prevent overfitting, and the target model before this position is regarded as the memorization-only, while the model after this position is considered as the overfitting. 
We then evaluate the attack performance of $\text{PFAMI}_{\textit{Met}}$, $\text{PFAMI}_{\textit{NNs}}$ and the best baselines ($\text{SecMI}_{\textit{Stat}}$ and $\text{SecMI}_{\textit{NNs}}$) across both memorization-only and overfitting models. The results are summarized in Fig.~\ref{fig:mia_traj}, where the attack performance of $\text{PFAMI}_{\textit{Met}}$ and $\text{PFAMI}_{\textit{NNs}}$ are more stable and robust across generative models with different memorization degrees. $\text{SecMI}_{\textit{Stat}}$ and $\text{SecMI}_{\textit{NNs}}$ can strike high attack performance after the target model is cursed with overfitting, but their performance is significantly degraded on memorization-only models.
These results further validate the effectiveness and generalizability of our proposed method in improving MIA by characterizing the memorization trace in generative models.

\subsection{How the Probabilistic Fluctuation Works}

\begin{figure}[t]
    \centering
    {
    \includegraphics[width=.22\textwidth]{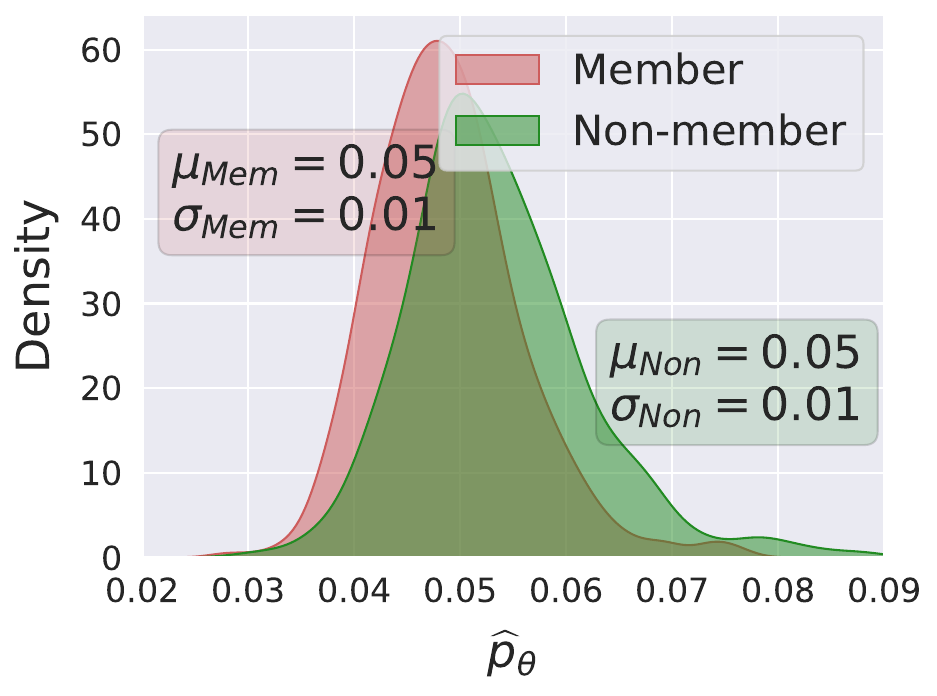}
    \includegraphics[width=.22\textwidth]{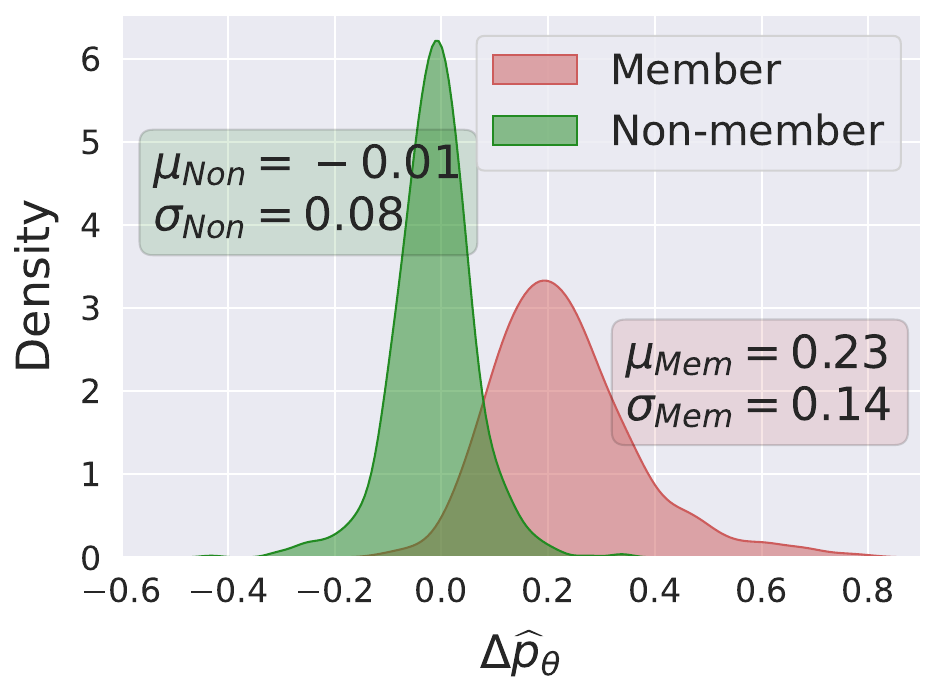}
    }\\
    \small {(a) Distributions of  $\widehat{p}_{\theta}$ (Left) and $\triangle \widehat{p}_{\theta}$ (Right) in DDPM.}\\
    \vspace{6pt}
    {
    \includegraphics[width=.22\textwidth]{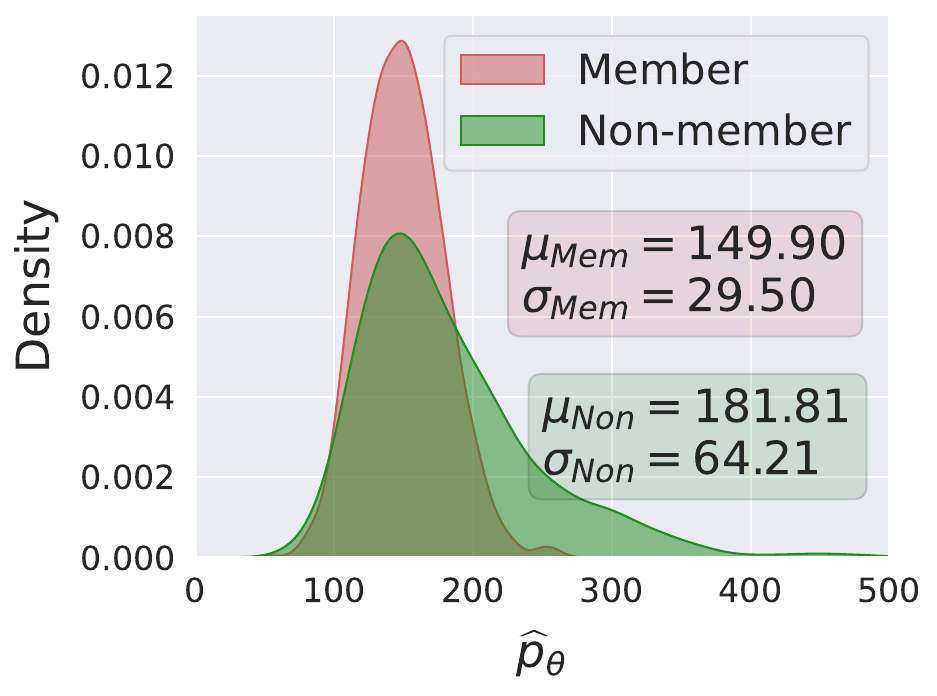}
    \includegraphics[width=.22\textwidth]{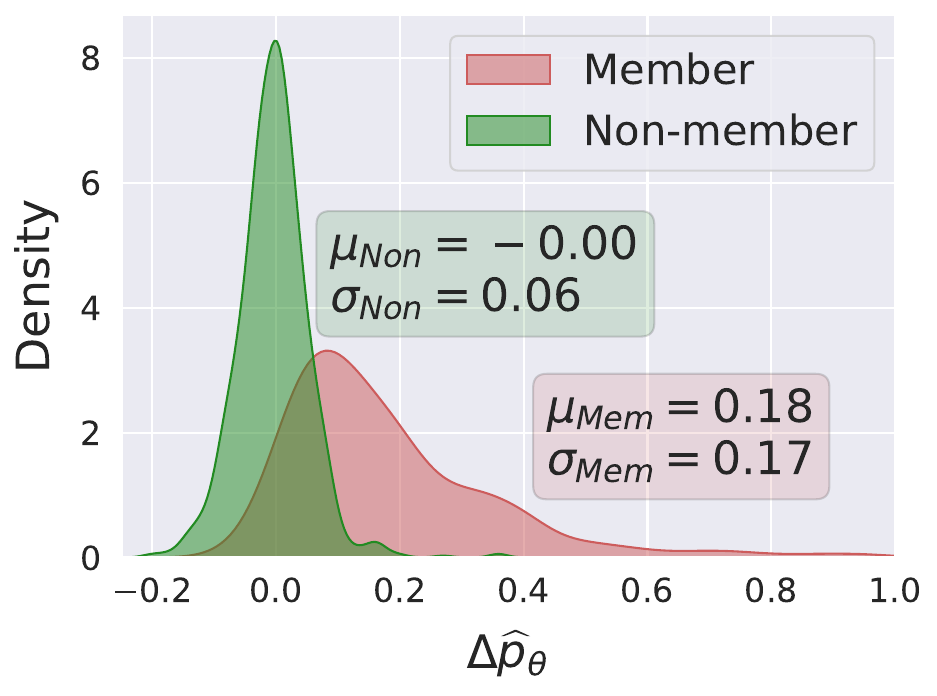}
    }\\
    \small {(b) Distributions of  $\widehat{p}_{\theta}$ (Left) and $\triangle \widehat{p}_{\theta}$ (Right) in VAE.}
    \caption{The discrimination of member and non-member records over approximate probability $\widehat{p}_{\theta}$ and approximate probabilistic fluctuation $\triangle \widehat{p}_{\theta} $ on DDPM and VAE.}
    \label{fig:distinguishability}
\end{figure}

We conduct detailed investigations about how our proposed probabilistic fluctuation works as a qualified extractor to distinguish member and non-member records. As exhibited in Fig.~\ref{fig:distinguishability}, we first visualize statistic distributions of the approximate probability $\widehat{p}_{\theta}$ and our proposed probabilistic fluctuation $\triangle \widehat{p}_{\theta}$ on member and non-member records. We found that just identifying the member records based on the approximate probability, as adopted by most existing works \cite{hilprecht2019monte,duan2023are}, is not reliable. Especially when the target generative model is not overfitting, the probability for both member and non-member data is very close, especially in the diffusion model, where their distributions almost overlap entirely. On the contrary, when using the probability fluctuation we designed, the obtained results have more substantial discriminative power. Specifically, due to the memorization effect, members are peak points in the probability distribution, and their neighbors tend to have lower probabilities. Hence, our designed probability fluctuation metric would be greater than zero. Conversely, non-members are typically located near inflection points in the generation probability distribution, resulting in a stable probability fluctuation around zero. These experimental results also demonstrate that $\text{PFAMI}_\textit{Met}$ can effectively enhance the discrimination between member and non-member records, making it easier to establish a threshold as a criterion for measuring record membership.

\begin{figure}[t]
    \centering
    {
    \includegraphics[width=.24\textwidth]{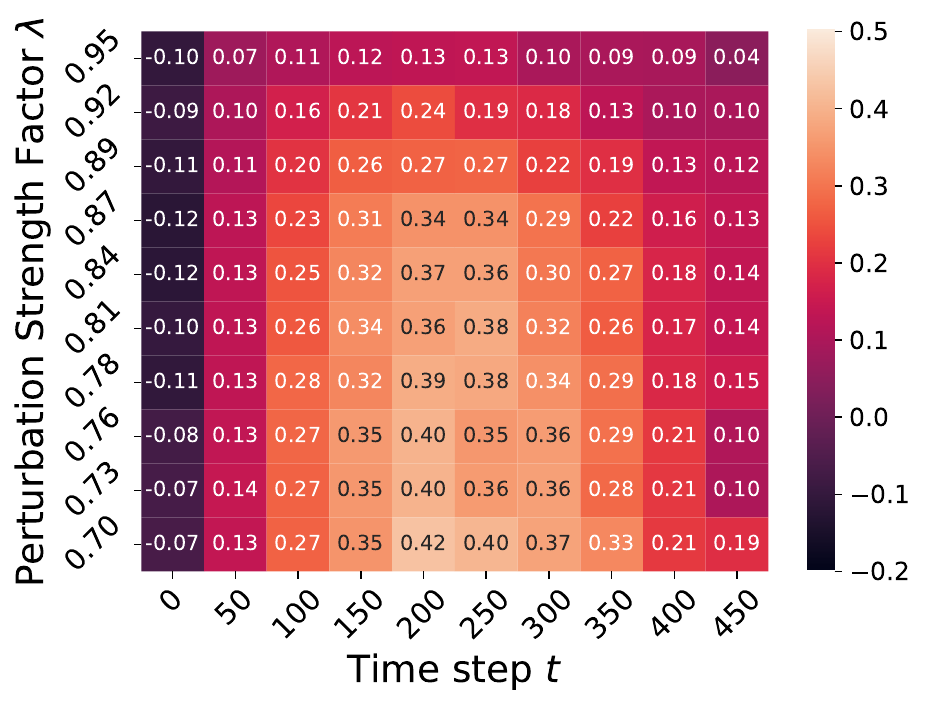}
    \includegraphics[width=.24\textwidth]{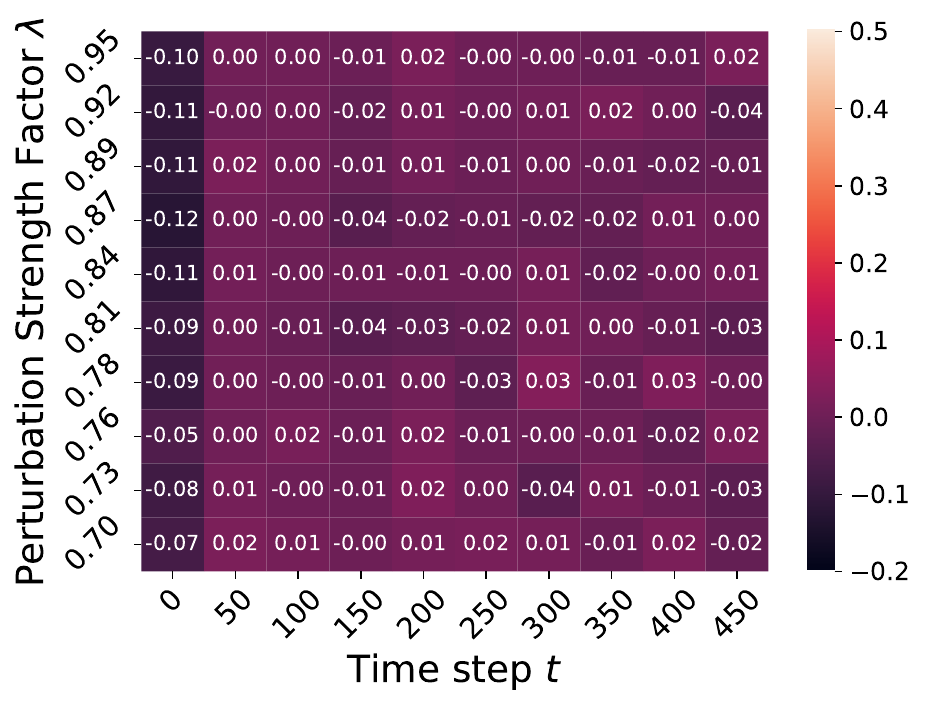}
    }\\
    \small {(a) $\boldsymbol{\triangle \widehat{p}_{\theta}}$ on member (Left) and non-member (Right).}\\
    \vspace{6pt}
    {
    \includegraphics[width=.23\textwidth]{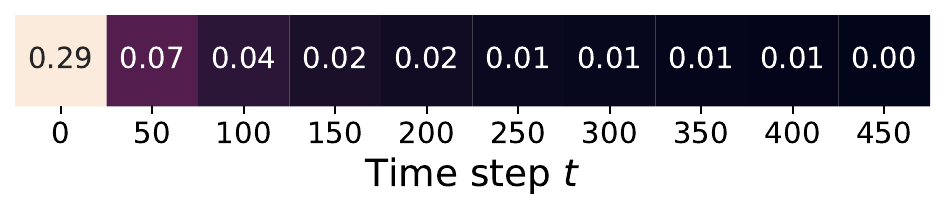}
    \includegraphics[width=.23\textwidth]{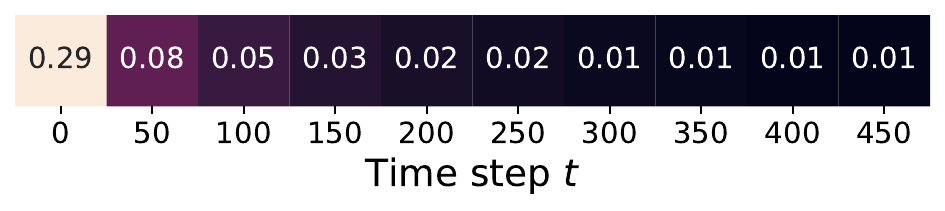}
    }\\
    \small {(b) $\boldsymbol{\widehat{p}_{\theta}}$ on member (Left) and non-member (Right).}
    \caption{The visualization of the probabilistic fluctuation $\boldsymbol{\triangle \widehat{p}_{\theta}}$  and the approximate probability $\boldsymbol{\widehat{p}_{\theta}}$ for member and non-member records on DDPM trained with Celeba-64.}
    \label{fig:heatmap}
\end{figure}

We also visualize the figure of probabilistic fluctuation $\boldsymbol{\triangle \widehat{p}_{\theta}}$ on diffusion models to help understand why the NNs-based approach can further improve the attack performance. As shown in Fig.~\ref{fig:heatmap}(a), we can observe an overall positive probability fluctuation of member records, except there are also apparent regular variations of probabilistic fluctuation over time step and perturbation strength. Consequently, this variation further benefits the NNs-based approach to extract the difference between member and non-member records. In contrast, as shown in Fig.~\ref{fig:heatmap}(b), the member and non-member records are indistinguishable only with the approximate probability on sampled time steps.

\subsection{Ablation Study}
We conduct an ablation study to investigate the performance gain provided by each module. Specifically, we respectively remove the variational probability assessment and neighbor record sampling modules as we introduced in Sec.~\ref{par:generation probability} and Sec.~\ref{par:perturbation}. Then we remove the NNs-based inference function proposed in Sec.~\ref{par:probabilistic fluctuation}. The results are presented in Table~\ref{tab:ablation}, where each module demonstrates significant improvement in performance gain, and the combination of all modules achieves the highest ASR of $\text{PFAMI}_{\textit{NNs}}$. The results also suggested that the approximate probability assessment is the basis of probabilistic fluctuation. Without this, it is difficult for our method to accurately estimate the fluctuation of generative probabilities, resulting in poor attack effectiveness.

\begin{table}
\centering
\tabcolsep=0.3em
\caption{Results of ablation study on the Celeba-64 dataset.}
\label{tab:ablation}
\resizebox{0.97\linewidth}{!}{%
\begin{tabular}{cccccc} 
\hline
\multirow{2}{*}{Methods}       & \multicolumn{2}{c}{DDPM} &  & \multicolumn{2}{c}{VAE}  \\ 
\cline{2-3}\cline{5-6}
                              & ASR$\uparrow$   & AUC$\uparrow$              &  & ASR$\uparrow$ & AUC$\uparrow$                \\ 
\hline\hline
$\text{PFAMI}_\textit{NNs}$   & 0.947 & 0.986            &  &  0.863  & 0.939                   \\ 
\hline
w/o Variational Probability Assessment  & 0.537 & 0.541            &  &  0.523   &     0.532               \\ 
\hline
w/o Neighbor Records Sampling & 0.694 & 0.781            &  &   0.677  &    0.747                \\
\hline
w/o NNs-based Inference Function   & 0.909 & 0.965            &  &  0.822  & 0.900                   \\ 
\hline
\end{tabular}
}
\end{table}


\subsection{Interpretation of Probabilistic Fluctuation as Directional Second Derivative} \label{par: interpretation}
In multivariable calculus, the second partial derivative test is generally utilized to authenticate whether a critical point is a local maxima, minima, or saddle point. From the inherent concept of this work, the member record is believed to be located on the local maximum of the probability distribution, where the Hessian matrix is negative definite, i.e., all the directional second derivatives are negative. However, it is intractable to determine whether all second-order directional derivatives of a sample $\boldsymbol{x}$ are negative. Moreover, we need only an approximate judgment in the region of local maxima, rather than strictly defined mathematical critical points. Therefore, we consider estimating the position of record $\boldsymbol{x}$ in the probability distribution by its expected value of second-order directional derivatives:
\begin{equation}\label{equ:second derivative}
   \mathbb{E}_{\boldsymbol{z}} \left( \partial^2_{uu} {p}_{\theta}\left(\boldsymbol{x}^{}\right) \right) = \mathbb{E}_{\boldsymbol{z}}\left(\boldsymbol{z}^{\top} H_p\left(\boldsymbol{x}^{}\right) \boldsymbol{z}\right),
\end{equation}
where  $H_p(\cdot)$ represents the hessian matrix of the probability function $p_\theta(\cdot)$ parameterized by the generative model $\theta$. This expression can be further approximated with the symmetric form:
\begin{equation} \label{equ: approx second}
    \boldsymbol{z}^{\top} H_p(\boldsymbol{x}^{}) \boldsymbol{z} \approx \frac{p_\theta(\boldsymbol{x}^{}+h \boldsymbol{z})+p_\theta(\boldsymbol{x}^{}-h \boldsymbol{z})-2 p_\theta(\boldsymbol{x}^{})}{h^2},
\end{equation}
where requires $h \to 0$, and $\boldsymbol{z}$ can be considered as "perturbation shift". Thus, $\boldsymbol{x}^{} \pm h \boldsymbol{z}$ can be regarded as neighbor text records of $\boldsymbol{x}^{}$ in the data distribution. From a statistical perspective, it is reasonable to assume that this "shift" is symmetric around the center of $\boldsymbol{x}$, then we further omit the factor $h$ and simplify Eq.~\ref{equ:second derivative} as follows:
\begin{equation}\label{equ:shift}
\begin{aligned}
        \mathbb{E}_{\boldsymbol{z}} \left( \partial^2_{uu} {p}_{\theta}\left(\boldsymbol{x}^{}\right) \right)
    & = \mathbb{E}_{\boldsymbol{z}} \left( p_\theta\left(\boldsymbol{x}^{}+\boldsymbol{z}\right) \right) - p_\theta\left(\boldsymbol{x}^{}\right) \\
    & = \mathbb{E}_{\widetilde{\boldsymbol{x}}^{} \sim q(\cdot \mid \boldsymbol{x}^{})} \left( p_\theta\left(\widetilde{\boldsymbol{x}}^{} \right) \right) - p_\theta\left(\boldsymbol{x}^{}\right),
\end{aligned}
\end{equation}
where $q(\cdot \mid \boldsymbol{x}^{})$ is a perturbation function that gives a distribution over $\widetilde{\boldsymbol{x}}^{}$, mildly shift the original image $\boldsymbol{x}^{}$ and maintain the outline and key features, which aligns with Eq.~(\ref{equ: approx second}) that requires the perturbation $h \to 0$.
Then, we can approximate Eq.~(\ref{equ:shift}) through Monte Carlo sampling:
\begin{equation}
    \begin{aligned}
        \mathbb{E}_{\boldsymbol{z}} \left( \partial^2_{uu} {p}_{\theta}\left(\boldsymbol{x}^{}\right) \right)
    & = \frac{1}{M}\sum_{j=i}^M {p}_{\theta}\left(\widetilde{\boldsymbol{x}}^{}_j\right) - {p}_{\theta}\left(\boldsymbol{x}^{}\right) \\
    & \propto - \frac{1}{M} \sum_{j=1}^M  \Delta {p}_{\theta} \left(\boldsymbol{x}^{}, \widetilde{\boldsymbol{x}}^{}_j\right),
    \end{aligned}
\end{equation}
which is exactly the overall probabilistic fluctuation measured in Eq.~(\ref{equ: probabilistic fluctuation}).

\subsection{How to Choose the Perturbation Mechanism}\label{app:perturbation}

\begin{table*}
\centering
\caption{Performance of $\text{PFAMI}_\textit{Met}$ on DDPM@Celeba-64 and VAE@Celeba-64 with different perturbation techniques.}
\label{tab:perturbation}
\resizebox{0.75\linewidth}{!}{%
\begin{tabular}{c|cccccccc} 
\hline
\multicolumn{2}{c}{\multirow{2}{*}{Perturbation Techniques}}         & \multicolumn{3}{c}{DDPM}                                                                                                          &                                      & \multicolumn{3}{c}{VAE}                                                                                                            \\ 
\cline{3-5}\cline{7-9}
\multicolumn{2}{c}{}                                                 & ASR                                       & AUC                                       & TPR @~$1\%$ FPR                           &                                      & ASR                                       & AUC                                       & TPR @~$1\%$~FPR                            \\ 
\hline\hline
\multirow{4}{*}{Color}    & Brightness                               & 0.855                                     & 0.920                                     & 0.291                                     &                                      & \textbf{0.843}                            & \textbf{0.919}                            & \textbf{0.237}                             \\
                          & Contrast                                 & 0.858                                     & 0.918                                     & 0.192                                     &                                      & 0.835                                     & 0.909                                     & 0.165                                      \\
                          & Saturation                               & 0.681                                     & 0.741                                     & 0.044                                     &                                      & 0.683                                     & 0.753                                     & 0.033                                      \\
                          & Hue                                      & 0.851                                     & 0.927                                     & 0.379                                     &                                      & 0.777                                     & 0.855                                     & 0.104                                      \\ 
\hline
\multirow{4}{*}{Geometry} & {\cellcolor[rgb]{0.855,0.855,0.855}}Crop & {\cellcolor[rgb]{0.855,0.855,0.855}}0.909 & {\cellcolor[rgb]{0.855,0.855,0.855}}0.965 & {\cellcolor[rgb]{0.855,0.855,0.855}}0.468 & {\cellcolor[rgb]{0.855,0.855,0.855}} & {\cellcolor[rgb]{0.855,0.855,0.855}}0.834 & {\cellcolor[rgb]{0.855,0.855,0.855}}0.912 & {\cellcolor[rgb]{0.855,0.855,0.855}}0.229  \\
                          & Rotation                                 & \textbf{0.917}                            & \textbf{0.968}                            & \textbf{0.574}                            &                                      & 0.806                                     & 0.879                                     & 0.119                                      \\
                          & Perspective                              & 0.836                                     & 0.911                                     & 0.167                                     &                                      & 0.730                                     & 0.799                                     & 0.049                                      \\
                          & Downsampling                             & 0.838                                     & 0.910                                     & 0.297                                     &                                      & 0.830                                     & 0.909                                     & 0.156                                      \\
\hline
\end{tabular}
}
\end{table*}

The perturbation mechanism is a key module in $\text{PFAMI}$ for characterizing the overall probabilistic fluctuation around the target record, which helps the attacker to sample representative neighbor records of the target record. Therefore, we investigate what perturbation mechanisms will be appropriate for $\text{PFAMI}$ to achieve better performances. Inspired by existing data augmentation techniques, we have studied two different types of data augmentation techniques, namely, color-based and geometry-based. They respectively involve various operations such as brightness, contrast, saturation, hue, as well as cropping, rotation, perspective, padding. We present several metrics to evaluate each perturbation mechanism. Except for ASR and AUC, we also consider TPR @~$1\%$ FPR, i.e., the TPR when FPR is  1\%. It can more accurately evaluate the attack performance when most mechanisms achieve a near-perfect AUC of approximately 0.9 and ASR scores. The results are shown in Table~\ref{tab:perturbation}, from which we can observe that the crop combines excellent performance and stability. Therefore, we adopt crop as the default perturbation mechanism in all experiments.
The results are presented in Table~\ref{tab:perturbation}, which illustrates that the crop demonstrates remarkable performance and robustness over two generative models. Consequently, the crop is selected as the default perturbation mechanism for all experimental trials.

\subsection{Impact of the Query Times}\label{par: query times}
\begin{figure}[t!]
    \centering
    \begin{tabular}{cc}
    {\includegraphics[width=.225\textwidth]{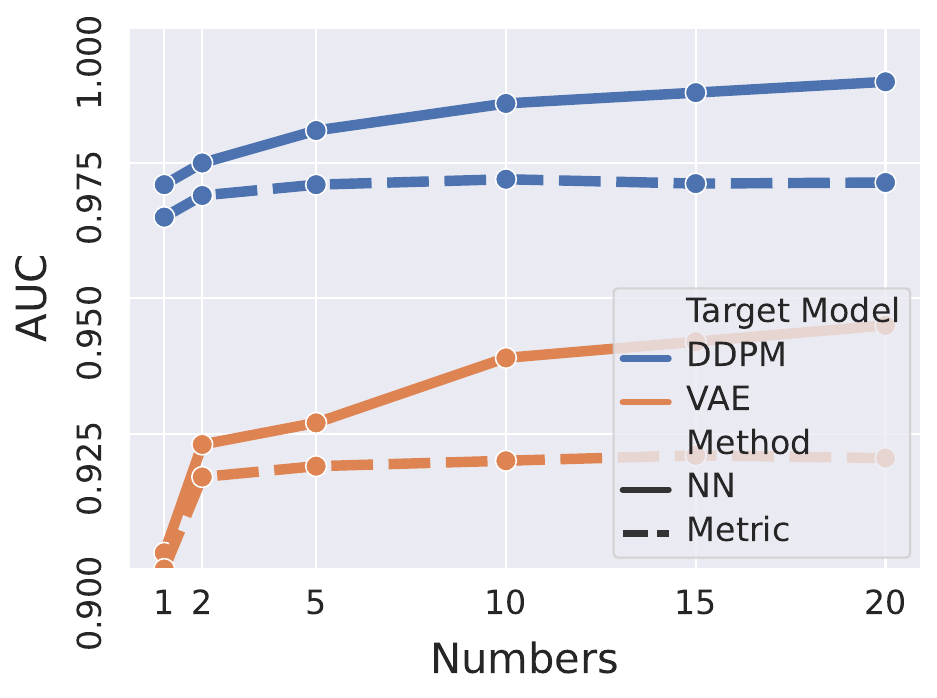}}&
    {\includegraphics[width=.225\textwidth]{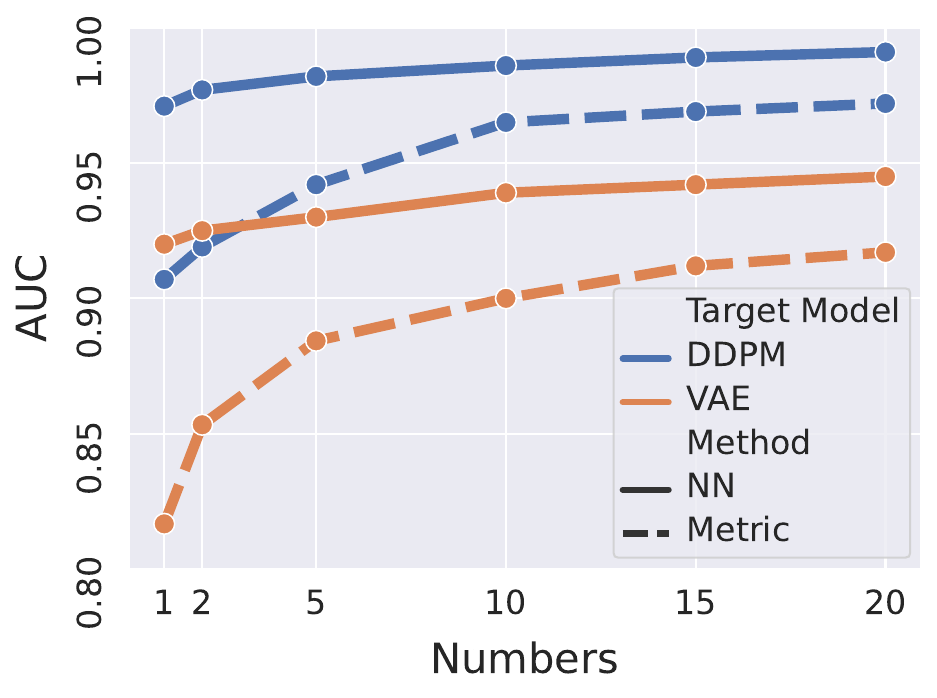}}\\
    \small{(a) \# Neighbor Record $M$}&
    \small{(b) \# Sampled Point $N$}
    \end{tabular}

    \caption{The attack performances against DDPM and VAE on Celeba-64 w.r.t the number of queries.}\label{fig:query times}
\end{figure}
We investigate the impact of query numbers on the performance of PFAMI attacks from two perspectives: the number of neighbor records $M$ and the number of sampled points $N$. Furthermore, we have conducted a detailed investigation into the variations in attack performance with respect to query numbers, considering different attack strategies and different architectures of generative models. Note that the default number of neighbor records $M$ is repetitively set to 1 and 10 in  $\text{PFAMI}_{\textit{Met}}$ and $\text{PFAMI}_{\textit{NNs}}$, and the default number of sampled points $N$ is set to 10. As the results presented in Fig.~\ref{fig:query times}, the attack performance of the NN-based strategy is exceptionally stable, due to the integration of probabilistic fluctuation features from multiple directions. On the other hand, the attack strategy based on statistical metric tends to rely more on the number of samples and generally achieve a good balance between performance and efficiency with around 10 sampled numbers.

\section{Ethical Statement and Broader Impacts}
In this paper, we propose a membership inference attack method, PFAMI, which can be maliciously utilized to infer the privacy of a specific individual whose data has been collected to train a probabilistic generative model. Indeed, we acknowledge that PFAMI can bring considerable privacy risk to existing generative models. Thus, to mitigate the potential abuse of this research, all experimental results are concluded over widely adopted public datasets, guaranteeing that each member record we extract has already been exposed to the public, thus avoiding any additional privacy breaches. In addition, we have made our code publicly available to facilitate further research in finding suitable defense solutions. Therefore, we believe our paper can encourage future studies to consider not only the generative capabilities of models but also the aspect of public data privacy and security.

\section{Conclusion}
In this article, we first point out that existing MIA algorithms largely rely on overfitting in generative models, which can be avoided by several regularization methods. Thus, the performance of these MIAs cannot be guaranteed. To mitigate this flaw, we opt for a more general phenomenon: memorization. Memorization is inevitable in deep learning models, and we have found that this phenomenon can be detected in probabilistic generative models by estimating the probabilistic fluctuations within the local scope of the target records. Therefore, we present a Probabilistic Fluctuation Assessing Membership Inference Attack (PFAMI) based on the distinct probabilistic fluctuation characteristics of members and non-members.
We conduct comprehensive experiments to evaluate PFAMI with various baselines on diffusion models and VAEs across different datasets. The results demonstrate that PFAMI maintains higher ASR and robustness across various scenarios than all baselines. We leave extending our attack framework to GAN as the future works, where we may incorporate the samples from the generator and the confidence values from the discriminator to design a better method.

\bibliographystyle{ieeetr}

\bibliography{reference}

\vfill

\end{document}